\pdfoutput=1
\documentclass[11pt]{article}

\usepackage[]{acl}

\usepackage{times}
\usepackage{latexsym}

\usepackage[T1]{fontenc}

\usepackage[utf8]{inputenc}

\usepackage{microtype}

\usepackage{inconsolata}
\usepackage{CJKutf8}
\usepackage{graphicx}
\usepackage{booktabs}

\usepackage{color, xcolor}

\usepackage{booktabs}
\usepackage{multicol}
\usepackage{multirow, makecell, caption}
\usepackage{colortbl}
\usepackage{tikz}
\usepackage{arydshln}
\usepackage{pgfplots}
\usepackage{amsmath}
\usepackage{amssymb}

\usepackage{tabularx}
\usepackage{enumerate}

\definecolor{green1}{rgb}{0.2, 0.45, 0.7}
\definecolor{green2}{rgb}{0.6, 0.68, 0.85}
\definecolor{green3}{rgb}{0.78, 0.83, 0.92}
\definecolor{green4}{rgb}{0.92, 0.93, 0.95}
\definecolor{green5}{rgb}{0.94, 0.97, 1.0}

\definecolor{green1}{rgb}{0.19, 0.75, 0.45}
\definecolor{green2}{rgb}{0.56, 0.85, 0.68}
\definecolor{green3}{rgb}{0.78, 0.92, 0.83}
\definecolor{green4}{rgb}{0.92, 0.95, 0.93}
\definecolor{green5}{rgb}{1, 1, 1}
\makeatletter
\def\adl@drawiv#1#2#3{%
        \hskip.5\tabcolsep
        \xleaders#3{#2.5\@tempdimb #1{1}#2.5\@tempdimb}%
                #2\z@ plus1fil minus1fil\relax
        \hskip.5\tabcolsep}
\newcommand{\cdashlinelr}[1]{%
  \noalign{\vskip\aboverulesep
           \global\let\@dashdrawstore\adl@draw
           \global\let\adl@draw\adl@drawiv}
  \cdashline{#1}
  \noalign{\global\let\adl@draw\@dashdrawstore
           \vskip\belowrulesep}}
\makeatother

\title{
QUILL: Quotation Generation Enhancement of Large Language Models}

\author{
 \textbf{Jin Xiao\textsuperscript{1}},
 \textbf{Bowei Zhang\textsuperscript{1}},
  \textbf{Qianyu He\textsuperscript{2}},
 \textbf{Jiaqing Liang\textsuperscript{1}\thanks{\enspace Corresponding author.}}
 \\
 \textbf{Feng Wei\textsuperscript{3}},
 \textbf{Jinglei Chen\textsuperscript{3}},
 \textbf{Zujie Liang\textsuperscript{3}},
  \textbf{Deqing Yang\textsuperscript{1}},
 \textbf{Yanghua Xiao\textsuperscript{2}}
 \\
 \textsuperscript{1}School of Data Science, Fudan University
 \\
 \textsuperscript{2}Shanghai Key Laboratory of Data Science, School of Computer Science, Fudan University
 \\
 \textsuperscript{3}MYbank, Ant Group
 \\
    \{jinxiao23, bwzhang24, qyhe21\}@m.fudan.edu.cn, \\
    \{liangjiaqing, yangdeqing, shawyh\}@fudan.edu.cn\\
    \{huodeng.wf, chenjinglei.cjl\}@mybank.cn\\
    \{jokieleung\}@outlook.com
}

\begin{document}
\thispagestyle{empty}
\maketitle

\begin{abstract}
While Large language models (LLMs) have become excellent writing assistants, they still struggle with quotation generation.
This is because they either hallucinate when providing factual quotations or fail to provide quotes that exceed human expectations.
To bridge the gap, we systematically study how to evaluate and improve LLMs' performance in quotation generation tasks.
We first establish a holistic and automatic evaluation system for quotation generation task, which consists of five criteria each with corresponding automatic metric.
To improve the LLMs' quotation generation abilities, we construct a bilingual knowledge base that is broad in scope and rich in dimensions, containing up to 32,022 quotes.
Moreover, guided by our critiria, we further design a quotation-specific metric to rerank the retrieved quotations from the knowledge base.
Extensive experiments show that our metrics strongly correlate with human preferences. 
Existing LLMs struggle to generate desired quotes, but our quotation knowledge base and reranking metric help narrow this gap.
Our dataset and code are publicly available at
https://github.com/GraceXiaoo/QUILL.
\end{abstract}

\section{Introduction}
\begin{figure}[t] 
    \centering
        \includegraphics[width=0.5\textwidth]{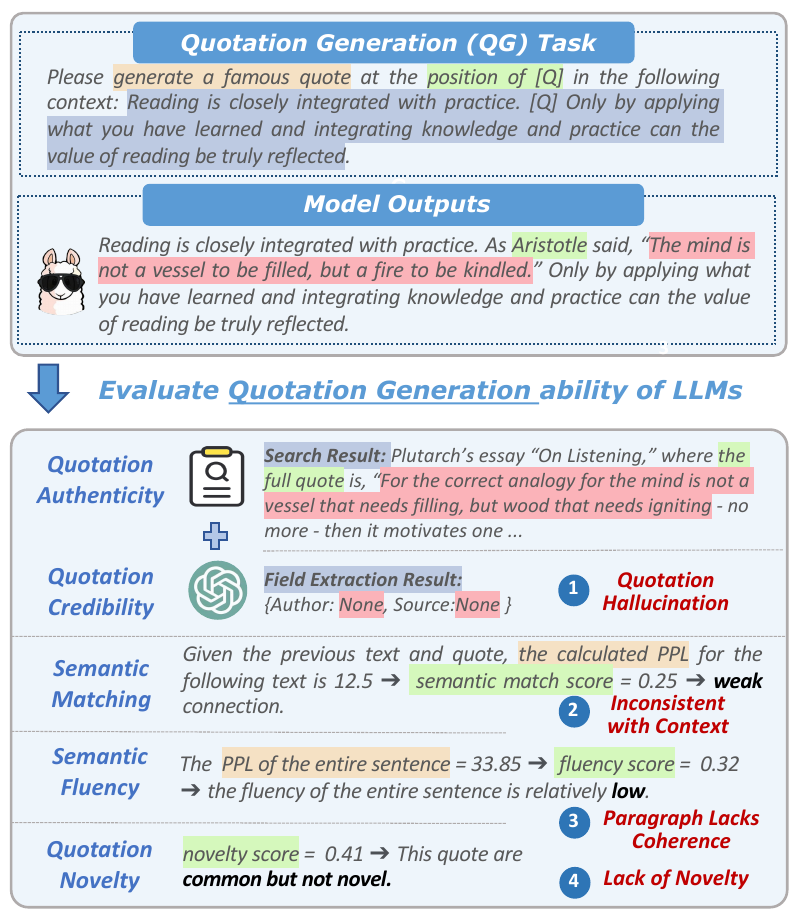}
    \caption{
    An example of prevalent issues in Quotation Generation (QR) by LLMs. In QR tasks, LLMs often fabricate sentences, leading to quotation hallucination. Additionally, the generated quotes may not align with the context, resulting in contextual inconsistency and semantic incoherence. Finally, the sentences produced by LLMs tend to be overly common, resulting in a lack of novelty in quotations.
    }
    \label{fig:010intro}
\end{figure}

\begin{figure*}[t] 
    \centering
            \includegraphics[width=1\textwidth]{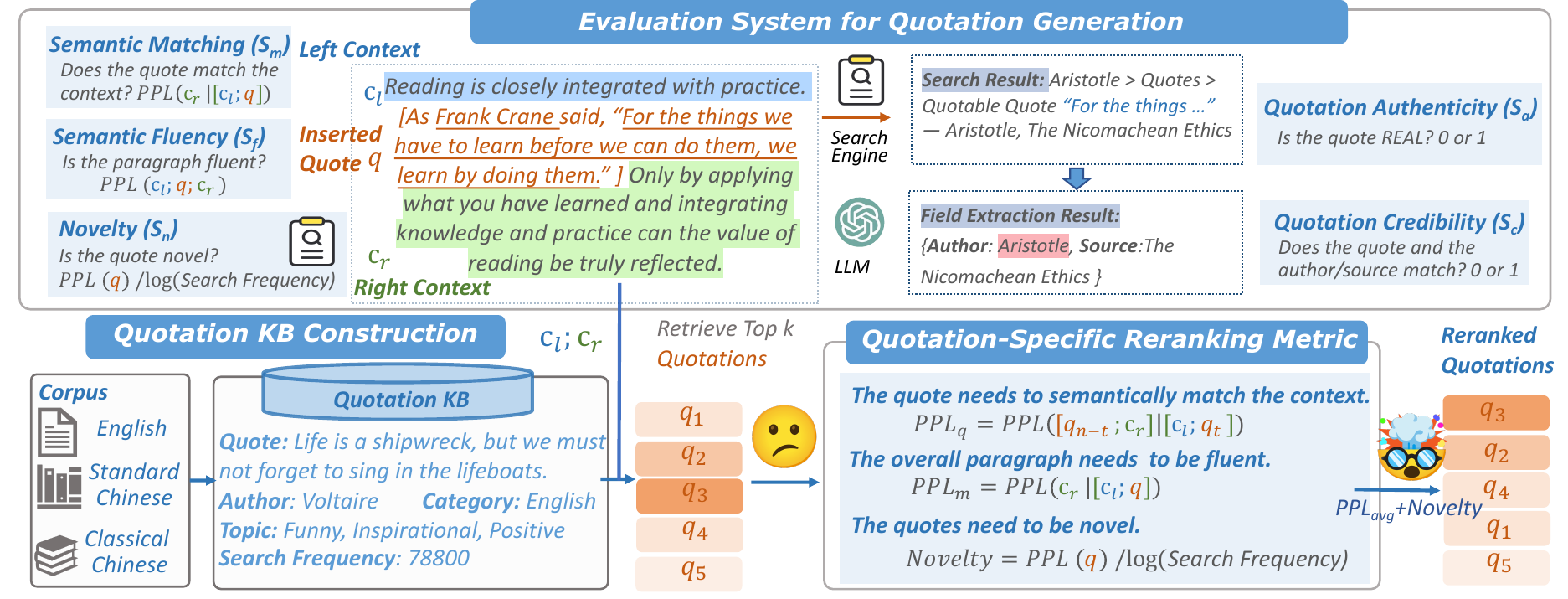}
    \caption{
    The framework for our Quotation Generation (QG) task research. We first establish an evaluation system with 5 evaluation criteria and automatic metrics, then build a quotation knowledge base covering multiple languages, topics and eras, and finally propose a quotation-specific reranking metric to rerank the quotations recalled in the RAG stage and improve the performance of QG tasks.
    }
    \label{fig:010method}
\end{figure*}


Famous quotations~\cite{Tan_Wan_Xiao_2015} are vital in academic and everyday communication. They lend authority to arguments and enhance persuasiveness, as they often stem from historically influential figures whose ideas have endured. 
Additionally, these quotations elevate the literary and artistic quality of a text, making discussions more engaging. They also facilitate comprehension of complex concepts, enabling readers to grasp ideas efficiently through concise expressions~\cite{vaswani2023attentionneed}.

The task of Quotation Generation (QG) seeks to produce suitable quotations to deepen the context in large language models (LLMs) ~\cite{anil2023palm,achiam2023gpt,touvron2023llama}. However, LLMs encounter significant challenges in this domain, as illustrated in Figure \ref{fig:010intro}. Primarily, the generated quotes frequently fail to correspond to genuine famous quotations and are often inaccurately attributed, a phenomenon termed "Quotation Halluciantion."~\cite{huang2023surveyhallucinationlargelanguage,bang2023multitaskmultilingualmultimodalevaluation,guerreiro2023hallucinationslargemultilingualtranslation} Additionally, these quotes don't align with the contextual meaning, resulting in a lack of coherence within the paragraph. Furthermore, LLMs exhibit a tendency to reproduce well-known quotes, which diminishes novelty and restricts creative expression. 

Although the issues of Quote Generation task are particularly problematic in LLMs, there is currently no effective solutions. Previous studies~\cite{qi-etal-2022-quoter} were based on representative pre-trained language models such as BERT~\cite{devlin2019bertpretrainingdeepbidirectional}, and it remains under-explored on the problem of quotation hallucination with LLMs. And there is currently no systematic and comprehensive benchmark to evaluate the quotation generation ability of LLMs.





To tackle these challenges, we introduce QUILL for \textbf{QU}otation Generat\textbf{I}on enhancement of \textbf{L}arge \textbf{L}anguage Models, a framework integrating an automatic evaluation system and an innovative and effective solution to improve quotation generation performance of LLMs.The framework of QUILL is shown in Fig.~\ref{fig:010method}. 
QUILL presents a comprehensive benchmark comprising 7 quotation domains and 16 real-world scenarios to evaluate large models' quotation generation abilities systematically, which consists of 5 highly interpretable and rigorous criteria with automatic evaluation metrics (Fig.~\ref{fig:010intro}): (1) \textbf{\textit{Quotation Authenticity}}: Confirm whether the quoted quotes are real quotes from famous people to prevent misquotations or fabrications. (2) \textbf{\textit{Quotation Credibility}}: Verify whether the quotation satisfies the author or source mentioned in the context (if any) to ensure the credibility of the quoted content. (3) \textbf{\textit{Semantic Matching}}: Evaluate whether the semantics of the quoted quote align with the context. (4) \textbf{\textit{Semantic Fluency}}: Evaluate the extent to which the cited quotation affects the fluency of the paragraph. (5) \textbf{\textit{Quotation Novelty}}: Evaluate the degree of uniqueness of the quote.

Additionally, based on the task's essential characteristics, we introduce an innovative Quotation-Specific Reranking Metric~\cite{karpukhin-etal-2020-dense,lewis2021retrievalaugmentedgenerationknowledgeintensivenlp,chern2023factoolfactualitydetectiongenerative} to improve model performance in QG tasks. To facilitate the task, we also established a comprehensive and high-quality knowledge database containing up to 32,022 quotes. This database spans both Chinese and English languages, various authors, different eras, and diverse topics, which ensures the wide applicability and generalization of our method.
To the best of our knowledge, our work is the first systematic investigation into the automatic evaluation and enhancement of quotation generation performance in LLMs. To summarize, our contributions are mainly four-fold: 
\begin{enumerate}
    \item We establish a holistic and automatic evaluation system for the quotation generation task, consisting of five highly interpretable and rigorous criteria, facilitating both human and automatic evaluation of this task.
    \item We construct a comprehensive and high-quality knowledge database containing up to 32,022 quotes, complete with authors or sources.
    \item We design a fine-grained quotation-specific metric to rerank the retrieved quotations from the knowledge base.
    \item We conduct extensive experiments to verify that our metrics are strongly correlate with human preference and significantly effective in both open-source and closed-source LLMs.
\end{enumerate}

\section{Related Work}

\subsection{Quotation}

Previous research about quotation mainly focused on quote recommendation~\cite{Tan_Wan_Xiao_2015}. The task of quote recommendation was initially proposed by~\cite{Tan_Wan_Xiao_2015}. They proposed a learning ranking framework for the task, which integrates 16 manually crafted features. \cite{10.1145/2911451.2914734} combined four different methods for recommending famous quotes, including matching granularity adjustment (a statistical context quote correlation prediction method), random forest, CNN, and LSTM. \cite{wang-etal-2020-continuity} utilized an encoder-decoder framework to generate speech responses based on separate modeling of dialogue history and current query. \cite{wang-etal-2021-quotation} used semantic matching to encode multi round dialogue histories using Transformer ~\cite{vaswani2023attentionneed} and GRU ~\cite{cho2014learningphraserepresentationsusing}, and encoded quotes using Transformer.
However, previous studies do not take into account the quotation generation capabilities of the large models themselves, nor did they propose a systematic and comprehensive evaluation system or benchmark to assess model performance in scenarios involving famous quotes.

\subsection{Hallucination}
In the field of NLP, hallucinations typically refer to a phenomenon where generated content appears meaningless or does not align with the provided source~\cite{filippova-2020-controlled,maynez-etal-2020-faithfulness}. 
To address the issue of hallucinations in language models, two primary methods have been proposed: (1) preventing hallucinations during the training and generation processes, and (2) reducing hallucinations after generation.~\cite{manakul2023selfcheckgptzeroresourceblackboxhallucination} introduced an alternative classification, dividing methods into black box and gray box approaches. Black box methods involve conducting factual checks without relying on external resources, either during or after generation. In contrast, gray box methods utilize external resources for validation.
Other techniques for alleviating hallucinations include reranking generated sample responses ~\cite{dale2022detectingmitigatinghallucinationsmachine} and improving beam search ~\cite{sridhar2023improvedbeamsearchhallucination}. Recent mitigation technologies have also shown promise in reducing hallucinations ~\cite{mündler2024selfcontradictoryhallucinationslargelanguage,pfeiffer2023mmt5modularmultilingualpretraining,chen2023purrefficientlyeditinglanguage,zhang2024knowledgealignmentproblembridging,agrawal2024languagemodelsknowtheyre}.
Although these methods have alleviated the quotation problem to a certain extent, they have not yet completely solved it, particularly in factual quotation and famous quotes.

\section{Background}



\subsection{Task Formulation}
\paragraph*{Quotation Generation}
Given a plain text \( c = [t_1, t_2, \dots, t_i, \dots, t_n] \), the goal of the \textit{Quotation Generation (QG)} task is to generate quotes for the specified insertion point \( i \). The left and right contexts, \( c_l \) and \( c_r \), are defined as \( c_l = [t_1, t_2, \dots, t_i] \) and \( c_r = [t_{i + 1}, \dots, t_n] \), respectively. In our work, we mainly focus on the ability of the model in quotation generation tasks.
\paragraph*{Quotation Recommendation}
In the \textit{Quotation Recommendation (QR)} task, given the context \( c = [t_1, t_2, \dots, t_i, \dots, t_n] \), the objective is to select the most suitable quote from the given set \( Q = \{q_1, \dots, q_{|Q|}\} \) to insert at position \( i \), where \( q_j \) represents the \( j\)-th quote.

\subsection{Preliminaries}
Perplexity (PPL) is a crucial metric in natural language processing, reflecting a model's predictive capability on text data and indicating the certainty of its next word prediction. Lower perplexity signifies greater confidence in the model's predictions, demonstrating a stronger ability to generate or understand language.
PPL of a language model given a sequence of words $w_1, w_2, \ldots, w_N$ is defined as:
\begin{equation}
\tiny
PPL \left(  P_r \mid P_l \right) = \exp\left(-\frac{1}{s} \sum_{i=t+1}^{N} \log P(w_i \mid w_1, \ldots, w_{i-1})\right)
\end{equation}
where \(P_{l}\) is the given left paragraph, $P_r$ is the following context needs to be calculated, $P(w_i \mid w_1, w_2, \ldots, w_{i-1})$ is the probability of the word $w_i$ given its left context, and $s$ is equal to N-t+1, which is the length of the sequence in the following paragraph.

\section{Evaluation System for QG}
The accuracy and rationality of quoting famous quotes are crucial, as they directly affect the credibility and rigor of the content. Therefore, we establish a holistic and automatic evaluation system for QG task evaluation in LLMs, containing five criteria and further design automatic metrics for each criterion (Fig.~\ref{fig:010intro}).

\paragraph{Criteria}Considering the nature of the quotation task itself, we design the following five criteria: (1) \textbf{\textit{Quotation Authenticity}}: Confirm whether the quoted quotes are real quotes from famous people to prevent misquotations or fabrications. (2) \textbf{\textit{Quotation Credibility}}: Verify whether the quotation satisfies the author or source mentioned in the context (if any) to ensure the credibility of the quoted content. (3) \textbf{\textit{Semantic Matching}}: Evaluate whether the semantics of the quoted quote align with the context. (4) \textbf{\textit{Semantic Fluency}}: Evaluate whether the quoted quote affects the fluency of the original text. (5) \textbf{\textit{Quotation Novelty}}: Evaluate the degree of uniqueness of the quote.

\paragraph{Evaluation Metrics}We propose automatic evaluation metrics for design standards, taking into account the essence of each metric. For any text containing the quote \( q \), the segment preceding the quote is termed the left context \( c_l \), while the segment following it is the right context \( c_r \). The combination of these segments forms the speech context \( c = [c_l; c_r] \).

\hspace{0.5em} \textbf{Quotation Authenticity.}
Authenticity of quotations is crucial as it ensures the reliability and credibility of information~\cite{kington2021identifying}. 
To verify the authenticity of the quoted celebrity quotes, 
we first search the quotation database for the information corresponding to the quote. If the database contains the information, we use the corresponding information to make a judgment. If not, we use different search engines (such as Google Scholar\footnote{\url{https://scholar.google.com/}} and Baidu Scholar\footnote{\url{https://xueshu.baidu.com/}}) to recall the corresponding search results.
Previous studies~\cite{han2024empiricalstudyinformationextraction} have shown that GPT-4o~\cite{chatgpt} has excellent simple extraction capabilities, and the extraction task based on this study only has two fields, author and source. 
Therefore, we use GPT-4o to extract the corresponding field information, and then compare the results of different search engines. If the field information is different, manual comparison is required.
For extraction details and validity of GPT-4o, please refer to Appendix~\ref{sec:GPT-4o}.
Finally, based on the extracted information, we verify whether the quote genuinely originates from the specific celebrity.
The final score is defined as follows:
\begin{equation}
\scriptsize
S_a = 
\begin{cases} 
1, & \text{if quote is real} \\
0, & \text{if not real}
\end{cases}
\end{equation}

\hspace{0.5em} \textbf{Quotation Credibility.}
Generally speaking, in the context of quoting, the source of the quote will be mentioned, such as the author, classic literature, or other sources. Ensuring consistency between the citation and the mentioned author or source is crucial for maintaining contextual coherence and information accuracy.
In order to confirm whether the citation meets the source restriction mentioned in the context, our study first extracts the source restriction of the context, and then compares and analyzes it with the extraction result of the previous indicator. If the source matches, the citation is marked as trustworthy, as shown in Fig.\ref{fig:010intro}.
The final score for quotation credibility is defined as follows:
\begin{equation}
\scriptsize
S_c = 
\begin{cases} 
1,&\text{if restriction matches} \\
0, & \text{if not match}
\end{cases}
\end{equation}

\hspace{0.5em} \textbf{Semantic Matching.}
Improper quotation may lead to misunderstandings or misinterpretations of the original meaning, thereby weakening the effectiveness and persuasiveness of the argument~\cite{sm}.
Perplexity is a common metric in NLP, used to assess a language model's predictive capability for text. 
Hence, we calculate the PPL of subsequent text based on a given prior text and quotation to evaluate the consistency between the quotation and its context. 
If the evaluation score is low, it implies that the citation aligns well with the following context in terms of semantics; otherwise, the rationality of the citation should be reconsidered.
The formula is as follows:

\begin{equation}
\scriptsize
PPL_m = PPL\left(c_r\mid[c_l;q] \right)
\label{equ:sm}
\end{equation}
where $c_{l}$ stands for the previous text, $q$ stands for the quotation, and $c_{r}$ stands for the following text.


To simplify computation, we normalize the PPL values to a range between 0 and 1. Given that a lower PPL indicates a higher degree of semantic alignment, we utilize an inverted Sigmoid function. The final calculation formula is as follows:

\begin{equation}
\scriptsize
S_m = \dfrac{1}{1+e^{k_m(PPL_m-\mu_m)}} * 100 \%
\end{equation}

where $\mu_m$ represents the mean value of $PPL_m$, which is 35.243, and $k_m$ is determined using an empirical formula, yielding a value of 0.053. 
See the Appendix~\ref{sigmoid} for the specific calculation details.

\hspace{0.5em} \textbf{Semantic Fluency.}
After quotation, it is necessary to ensure that the entire context is fluent and coherent to maintain semantic consistency and logical integrity~\cite{10.1093/arclin/acaa109}.
This study calculates the PPL of the entire context to measure the textual fluency of the overall context after inserting quotations.
Lower perplexity indicates smoother overall contextual semantics.
The calculation formula for semantic fluency is as follows:

\begin{equation}
\scriptsize
PPL_f = PPL_q \left([c_l,q,c_r]\mid \cdot \right)
\end{equation}
where $c_{l}$ stands for the previous text, $q$ stands for the quotation, and $c_{r}$ stands for the following text. 

Similarly, for normalizing the PPL values into a range from 0 to 1, the final score for semantic fluency is designed as follows:
\begin{equation}
\scriptsize
S_f = \dfrac{1}{1+e^{k_f(PPL_f-\mu_f)}} * 100 \%
\end{equation}
where $\mu_f$ represents the mean value of $PPL_f$, which is 16.470, and $k_f$ is determined using an empirical formula, yielding a value of 0.500.

\hspace{0.5em} \textbf{Quotation Novelty.}
Integrating novel quotations into established ideas enhances originality and personalizes the expression within academic discourse~\cite{savov2021measuring}. To evaluate the extent to which the quote introduces new ideas or unique perspectives to the original context, we utilize the Bing\footnote{\url{https://www.bing.com/}}search engine to determine the number of Search Frequency corresponding to each quotation, applying a log10 transformation to quantify quotation popularity.
In addition, to mitigate potential biases in search results, we also incorporate the quoted PPL value for supplementation. As a lower PPL indicates a higher frequency of text occurrence, it is inversely correlated with search frequency.
Therefore, the formula is as follows:

\begin{equation}
\scriptsize
\text{Novelty} =  \frac{PPL(q\mid \cdot)}{log_{10}(\text{Search Frequency})} 
\label{equ:novelty}
\end{equation}
where Search Frequency indicates the number of search results obtained by searching the quotation in the Bing search engine.
In order to map the PPL value to a range of 0 to 1, since higher novelty means higher score, the positive sigmoid function is adopted here, and the final score is as follows:
\begin{equation}
\scriptsize
S_n = \dfrac{1}{1+e^{-k_n(Novelty-\mu_n)}} * 100 \%
\end{equation}
where $\mu_n$ represents the mean value of Novelty, which is 10.67, and $k_n$ is determined using an empirical formula, yielding a value of 0.253.

\section{Quotation Knowledge Base}



\subsection{Dataset Construction}
In order to alleviate the problem of famous quote hallucination in LLMs, we develop a comprehensive bilingual and multi-topic quotation corpus designed to enhance retrieval quotation tasks during the RAG stage , as shown in Tab.~\ref{tab:data}. This corpus is structured into three distinct components: the English, the Standard Chinese, and the Classical Chinese. To improve the application scope and practical value of the corpus, we ensure comprehensive coverage of both common and specialized fields and also implement stringent quality control measures. Each quote is manually reviewed to ensure accuracy and relevance.

\newcolumntype{b}{>{\columncolor{blue!10}}r}
\newcolumntype{d}{>{\columncolor{brown!10}}r}
\newcolumntype{q}{>{\columncolor{Green!10}}r}

\setlength\tabcolsep{1pt}
\begin{table}[h]
    \centering
    \tiny
    \resizebox{0.45\textwidth}{!}{
        \begin{tabular}{lcccc}
            \toprule
            \textbf{Category} & \textbf{Samples} & \textbf{AvgLen} & \textbf{AvgSearchFreq} & \textbf{AvgNovelty} \\
            \midrule
            English            & 16,393 & 16         & 2,823,499 & 6.8 \\
            Standard Chinese   & 7,519  & 42         & 150,011   & 6.3 \\
            Classical Chinese  & 8,110  & 14         & 19,017    & 5.0 \\
            \midrule
            Total              & 32,022 & 24         & 997,509   & 6.0 \\
            \bottomrule 
        \end{tabular}
    }
    \caption{The statistics of our knowledge base. For each category, the \textit{AvgLen}, \textit{AvgSearchFreq} and \textit{AvgNovelty} denote the average of the length, the frequency of Bing Search engine, and the value of Quotation Novelty respectively.}
    \label{tab:data}
\end{table}

\paragraph{English Corpus}  To construct the English quotation corpus, we extract approximately 27,260 quotes covering different topics from the \textit{BrainyQuote}\footnote{\url{https://www.brainyquote.com/}}, \textit{A-ZQuote}\footnote{\url{https://www.azquotes.com/}} and \textit{Goodreads}\footnote{\url{https://www.goodreads.com/}} websites, categorizing them by topic and author.
\paragraph{Classical Chinese Corpus}  Considering the representativeness and novelty of the Chinese corpus, we first collect some famous citations from \textit{Gushiwen}\footnote{\url{https://www.gushiwen.cn/}}. Subsequently, given the limited number of citations, we utilize LLM to conduct a meaningful selection of the collected poems from \textit{BaiduHanyu}. For instance, the seven-character quatrains in Tang poetry can be divided into two citations. Furthermore, to enhance the generalization of themes, we employ LLM to summarize the topics of  the quotes. Finally , we collect over 9,233 citations with its poems, author and topics, including various genres such as Tang poetry and Song lyrics.
\paragraph{Standard Chinese Corpus} Regarding the Standard Chinese quotation corpus, we gather 13,453 quotes from the \textit{Guzimi}\footnote{\url{https://www.juzimi.com.cn/mingyan/}} and \textit{Mingyancidian}\footnote{\url{http://mingyan.juzicidian.com}} websites, similarly categorized by topic and author.

\paragraph{Dataset Evolution}  For those collected from diverse websites, the corpus have two limitations: (1) Semantic redundancy: the semantics of different quotations are too similar, especially when a long quotation includes a shorter one. (2) Lengthy quotations: some quotations are excessively long. Hence, We first utilized the Jaccard Similarity coefficient to address the issue of semantic redundancy. Then we set a restriction on the length of the citations and remove the extreme values based on the quotation ppl metric. 
Additionally, to facilitate the subsequent rerank stage of retrieval-augmented generation (RAG), we also pre-calculate the novelty of the quotations in the database. 
The specific calculation is detailed in Equation (\ref{equ:novelty}).
Finally, we obtain a higher-quality corpus exceeding 32,022 entries. The statistics of our knowledge dataset is as show in the Tab.~\ref{tab:data}.

\subsection{Dataset Statistics}
In this part, we compare the statistics of our dataset with existing quotation-related resources, as shown in Tab.\ref{tab:resource}.
In contrast, our dataset is the first to consider quotation novelty, encompassing a wide range of topics and numerous authors, while recording and annotating their sources. Additionally, we have expanded the scale of the quotation dataset, thereby broadening its application scenarios and significance.

\subsection{Quality Assessment by Human}

After constructing the dataset, we manually check its quality. For each component, we randomly select 100 quotes and engage three annotators to verify their validity. The annotators use search engines~\footnote{\url{https://www.bing.com/}} to locate references and evaluate both the authenticity of the quotes and the accuracy of their attributed authors and sources. Only quotes that satisfy both criteria are included in the final dataset.
The final results are determined through a majority voting process. 
In the English, Standard Chinese and Classical Chinese components, 99, 97 and 98 quotations respectively met the established criteria.
These results confirm the high quality of the dataset, which is derived from trustworthy sources such as published books and reputable citation websites.

\newcolumntype{b}{>{\columncolor{blue!10}}r}
\newcolumntype{d}{>{\columncolor{brown!10}}r}
\newcolumntype{q}{>{\columncolor{Green!10}}r}

\setlength\tabcolsep{1pt}
\begin{table}[h]
    \centering
    \tiny
    \resizebox{0.48\textwidth}{!}{
    \begin{tabular}{lccccc}
    \toprule
   \textbf{Resource} & \textbf{Size} & \textbf{Topic} &  \textbf{Author} & \textbf{Multilingual} &\textbf{Novelty}\\
   \midrule
   LRQW ~\cite{Tan2005} &3,158&822&762&N&N\\
   QRDW ~\cite{Ahn2016}&1,200&-&-&N&N\\
   QuoteR ~\cite{qi2022}&13,550&-&-&Y&N\\
   \midrule
   Ours&32,022&2,301&9,708&Y&Y\\
   \bottomrule 
    \end{tabular}
       }
    \caption{The statistics of our dataset with existing quotation-related resources. Multilingual refers to the inclusion of two or more languages, Y denotes Yes, and N denotes No.}
    \label{tab:resource}
\end{table}



\section{Quotation-specific Reranking Metric}

In our study we introduce a fine-grained and end-to-end RAG solution to improving model performance in quotation tasks through introducing a straightforward and interpretable quotation-specific rerank metric to select the optimal quotation.

When the user inputs the context to be inserted, we use semantic similarity to recall the top k most relevant quotes from the knowledge database.
However, while similarity assesses the semantic relevance between the quotation and the context, the QG task necessitates a more comprehensive approach. It requires not only that the semantics of the quote align with the context but also that the paragraph maintains fluency and incorporates novel citations.
To enhance the performance of LLMs in QG, we propose three evaluative sub-indicators as shown in Fig.~\ref{fig:010method}:


\paragraph{Quotation Matching}


Quotation matching emphasizes the completion of the quotation itself and its alignment with the subsequent text~\cite{maclaughlin-smith-2021-content}. This is accomplished by calculating the PPL of the remaining portion of the quotation, given the preceding text and the initial k characters of the quotation. 
Generally, lower PPL values suggest that the model produces more accurate and coherent quotations. 
The specific calculation formula is as follows:
\begin{equation}
\scriptsize
PPL_q = PPL \left([q_{n-t};c_r]\mid[c_l;q_t] \right)
\label{equ:qm}
\end{equation}
where $n$ represents the length of the quote, $q_t$ represents the first $t$ characters of the quote, $q_{n-t}$ represents the remaining $n-t$ characters of the quote.


\paragraph{Semantic Matching}
Semantic matching is concerned with ensuring semantic consistency and logical coherence within the context. This is achieved by predicting the PPL of the subsequent text, given the preceding text and the entire quote. 
A lower PPL value means that the quotation is more semantically consistent with the following context.
The calculation formula is as Equation (\ref{equ:sm}).

\begin{figure}[t] 
    \centering
        \includegraphics[width=0.48\textwidth]{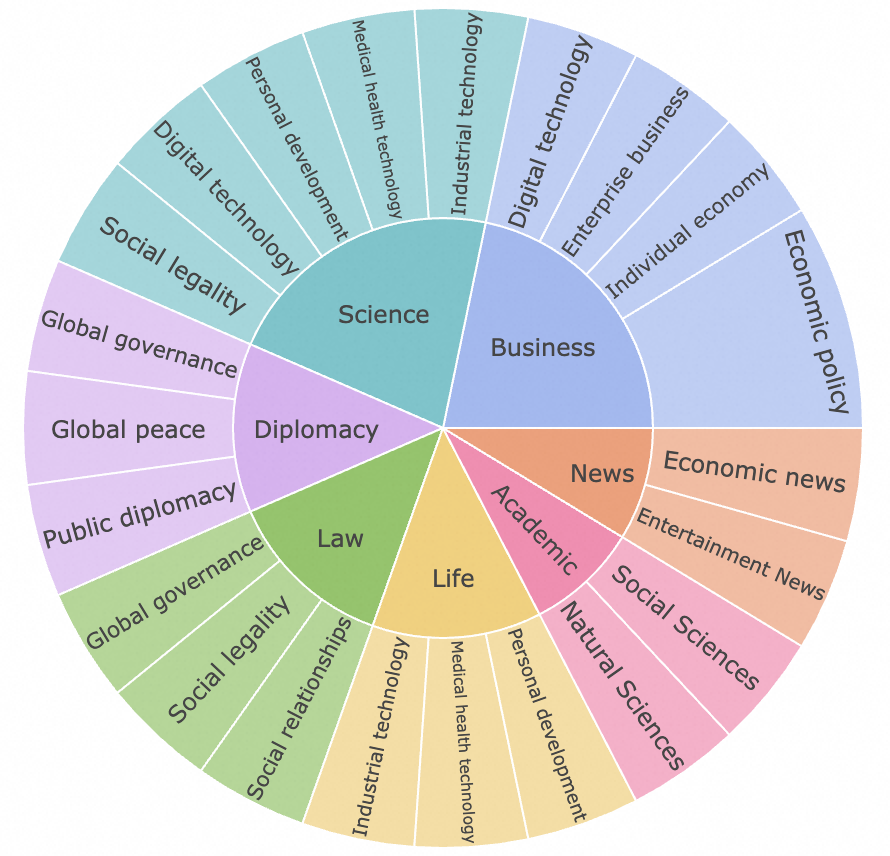}
    \caption{7 common categories and 21 scenarios 
 details of the evaluation dataset.}
    \label{fig:dataset}
\end{figure}
\paragraph{Novelty}
The Novelty metric evaluates the originality of generated quotations. By avoiding repetition and clichés, this metric ensures that content remains fresh and engaging, providing unique perspectives across various contexts.
The specific calculation formula is as Equation (\ref{equ:novelty}).

To integrate the advantages of the three indicators, we employ a weighted average method, utilizing it as our final quotation-specific rerank metric. This comprehensive indicator seeks to balance semantic matching, fluency, and novelty, thereby enhancing the overall quality of model-generated citations.
Finally, after the rerank stage, we select the top-1 quote including author or source information, and add it to the prompt. Then, the model inserts and rewrites quotes dynamically in the context, and ultimately outputs the results we need.

\section{Experiments}
\newcolumntype{b}{>{\columncolor{blue!10}}r}
\newcolumntype{d}{>{\columncolor{brown!10}}r}
\newcolumntype{q}{>{\columncolor{Green!10}}r}

\setlength\tabcolsep{1pt}
\begin{table}[t]
\centering
  \large
    \resizebox{0.45\textwidth}{!}{
    \begin{tabular}{lcccccd}
    \toprule
    \textbf{Model}
    &\textbf{$S_a$} & \textbf{$S_c$}& \textbf{$S_m$} & \textbf{$S_f$}  & 
    \textbf{$S_n$}  &\textbf{$Avg$}  \\
    \midrule
    \multicolumn{7}{l}{\textbf{\textit{Chinese-oriented Models}}} \\
    \midrule
ChatGLM3-6B & 0.56 & 0.20 & 0.72 & 0.73 & 0.71 & 0.58\\
Qwen1.5-7B-Chat & 0.63 & 0.15 & 0.72 & 0.68 & 0.71 & 0.58\\
Qwen1.5-14B-Chat & 0.66 & 0.16 & 0.72 & 0.69 & 0.73 & 0.60 \\
Qwen1.5-72B-Chat & 0.72 & 0.16 & 0.71 & 0.71 & 0.67 & 0.60  \\
Deepseek-R1 & 0.70 &\underline{0.39}&0.72&\textbf{0.76}&0.49&0.61\\
    \midrule
    \multicolumn{7}{l}{\textbf{\textit{English-oriented Models}}} \\
    \midrule
    Mixture-7B-v0.2 & 0.77 & 0.08 & 0.70 & 0.74 & 0.55 & 0.57 \\
    Llama2-7B-Chat-hf & 0.46 & 0.15 & 0.73 & 0.73 & 0.74 & 0.56\\
    Llama2-13B-Chat-hf & 0.48 & 0.15 & \underline{0.74} & 0.72 & \underline{0.74} & 0.56  \\
    Llama2-70B-Chat-hf & 0.60 & 0.11 & 0.69 & 0.67 & 0.62 & 0.55 \\
    \midrule
    \multicolumn{7}{l}{\textbf{\textit{Close-source Models}}} \\
    \midrule
    GPT-3.5-turbo & 0.62 & 0.11 & 0.71 & 0.72 & 0.62 & 0.56\\
    GPT-4o &  \underline{0.79} & 0.23 & 0.71 & 0.74 & 0.58  & \underline{0.61}\\
    \midrule
    Ours& \textbf{1.00} & \textbf{1.00} & \textbf{0.75} & \underline{0.75} & \textbf{0.81} & \textbf{0.86}\\
    \bottomrule 
    \end{tabular}
    }
    \caption{Comparison of performance of various models on our evaluation system for QG tasks.}
    \label{tab:result1}
\end{table}

In this section, we conduct experiments to verify the effectiveness of our method and metrics.

\subsection{Experiment Setup}
\paragraph{Evaluation Dataset}In constructing the evaluation dataset, our study select 7 common categories: economy, diplomacy, journalism, academia, law, technology, and life. Additionally, 21 frequently cited scenarios are chosen to encompass various aspects of the knowledge system, as show in Fig.~\ref{fig:dataset}. To enhance the dataset's diversity, standard Chinese, classical Chinese, and English texts are also included.
Initially, we gather quotes from each scenario to ensure diversity, richness, and relevance to the selected fields. 
After collecting these quotes, they are used as keywords to search on major search engines like Google\footnote{\url{https://www.google.com/}}, Bing\footnote{\url{http://www.bing.com}}, and Baidu\footnote{\url{https://www.baidu.com/}}. Then articles containing these quotes are identified, and the relevant context is extracted. To guarantee the dataset's quality, we perform preprocessing and cleaning, which involved removing duplicates, correcting errors, and eliminating ambiguities.
Then, we conduct manual sampling and validation to evaluate and ensure the dataset's quality and usability.
Finally, we obtain the evaluation dataset that comprises 600 context-quote pairs.
\paragraph*{Models} 
We evaluate 9 models ranging from their model sizes and structures, which fall into three categories: Chinese-oriented models, English-oriented models, and Close-source models.


\paragraph*{Models for PPL Calculation}
We employ two advanced language models, Qwen2-7B~\cite{bai2023qwen} and Llama3-8B~\cite{touvron2023llama}. These models are used to compute the PPL of the context given the preceding text. Subsequently, the average PPL values calculated by the two models are taken as final PPL values, which balances the judgments of the two models and reduces potential bias introduced by any single model.
Since larger models tend to produce lower PPL for the same text, we recommend using the same PPL calculation models in this study when evaluating QG tasks.


\newcolumntype{b}{>{\columncolor{blue!10}}r}
\newcolumntype{d}{>{\columncolor{brown!10}}r}
\newcolumntype{q}{>{\columncolor{Green!10}}r}

\setlength\tabcolsep{1pt}
\begin{table}[t]
    \centering
    \tiny
    \resizebox{0.5\textwidth}{!}{
    \begin{tabular}{lccccc}
    \toprule
   \textbf{Method} & \textbf{HR@1} & \textbf{HR@3} &  \textbf{nDCG@1} & \textbf{nDCG@3} & \textbf{MRR} \\
   \midrule
   Vanilla & 0.13 & 0.43 & 0.50 & 0.72 & 0.35 \\
   \midrule
    \multicolumn{6}{l}{\textbf{\textit{Supervised}}} \\
    \midrule
    BM25 & 0.19 & 0.50 & 0.54 & 0.71 & 0.39 \\
    monoT5 (3B) & 0.31 & 0.61 & 0.65 & 0.77 & 0.48 \\
    \midrule
    \multicolumn{6}{l}{\textbf{\textit{Unsupervised}}} \\
    \midrule
    UPR (FLAN-T5-XL) & 0.31 & 0.52 & 0.63 & 0.74 & 0.46 \\
    bge-reranker-large & 0.32 & 0.55 & 0.71 & 0.82 & 0.47 \\
    \midrule
    \multicolumn{6}{l}{\textbf{\textit{LLM API (Permutation Generation)}}} \\
    \midrule
    GPT-3.5-turbo & 0.33 & 0.61 & 0.72 & \underline{0.84} & 0.50 \\
    GPT-4o & 0.43 & 0.63 & \underline{0.74} & \textbf{0.88} & 0.55 \\
    \midrule
    \multicolumn{6}{l}{\textbf{\textit{Quotation-specific Reranking Metric}}} \\
    \midrule
    PPL\(_\text{q}\) & 0.45 & \underline{0.66} & 0.71 & 0.83 & 0.57 \\
    PPL\(_\text{m}\) & 0.34 & 0.60 & 0.64 & 0.77 & 0.50 \\
    PPL\(_\text{avg}\) & 0.33 & 0.60 & 0.64 & 0.76 & 0.50 \\
    PPL\(_\text{q}\) + Novelty & 0.34 & 0.58 & 0.63 & 0.73 & 0.50 \\
    PPL\(_\text{m}\) + Novelty & \underline{0.46} & 0.65 & 0.70 & 0.78 & \underline{0.57} \\
    PPL\(_\text{avg}\) + Novelty (ours) & \textbf{0.49} & \textbf{0.67} & \textbf{0.74} & 0.79 & \textbf{0.60} \\
   \bottomrule 
    \end{tabular}
    }
    \caption{Performance of different rerank metrics on Hit@1, Hit@3, nDCG@1, nDCG@3 and MRR.
    $PPL_q$, $PPL_m$ and Novelty are as defined in Section 6, and $PPL_{avg}$ is the average of $PPL_q$ and $PPL_m$.
    Best performing reranker method are marked bold.}
    \label{tab:result2}
\end{table}

\begin{figure}[t] 
    \centering
            \includegraphics[width=0.5\textwidth]{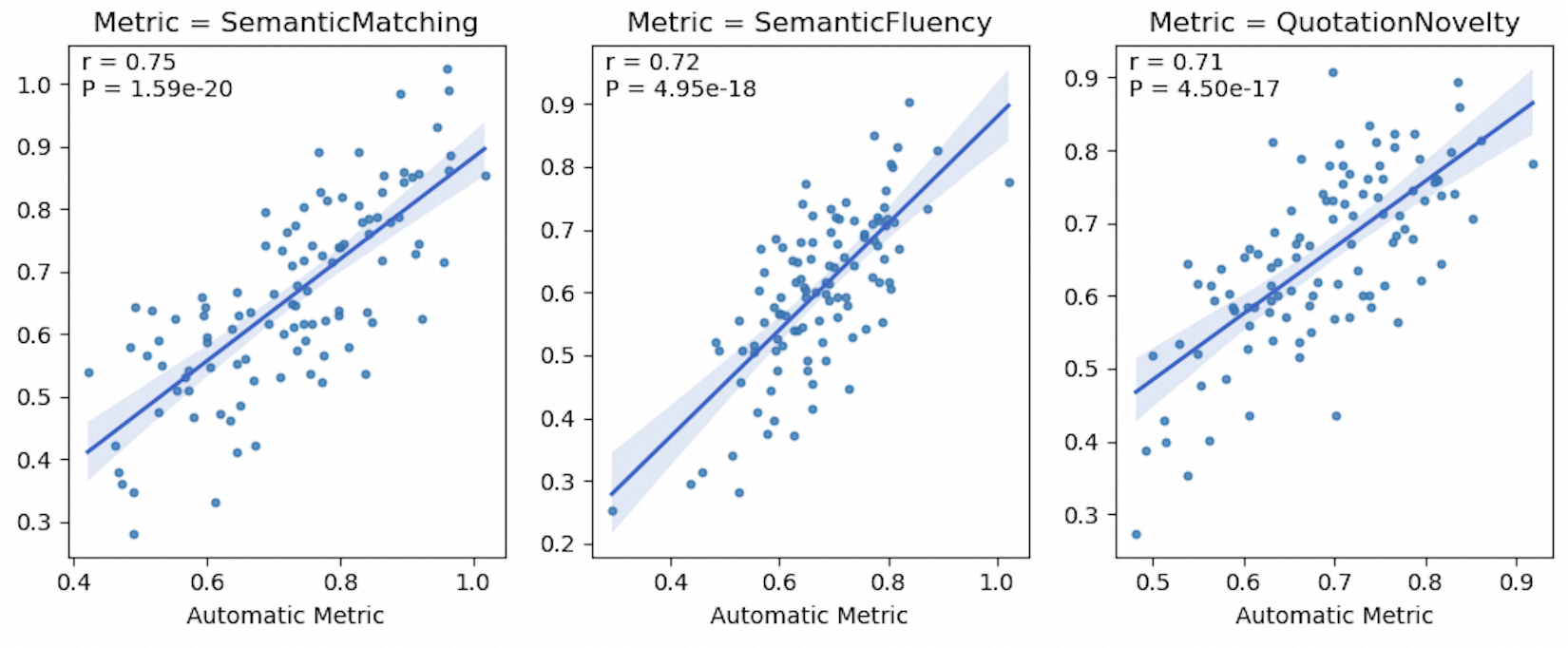}
    \caption{Correlation between our automatic evaluation metrics and human ratings. To avoid overlapping points, random jitters
sampled from $N (0, {0.05}^2)$ were added to human ratings after fitting the regression.}
    \label{fig:070relation}
\end{figure}

\newcolumntype{b}{>{\columncolor{blue!10}}r}
\newcolumntype{d}{>{\columncolor{brown!10}}r}
\newcolumntype{q}{>{\columncolor{Green!10}}r}

\setlength\tabcolsep{1pt}
\begin{table*}[!htb]
    \centering
    \tiny
    \resizebox{1\textwidth}{!}{
    \begin{tabular}{lcccccd cccccb cccccd cccccb}
    \toprule
    & \multicolumn{6}{c}{\textbf{Naive-0-Shot}} 
    & \multicolumn{6}{c}{\textbf{Naive-1-Shot}}
    & \multicolumn{6}{c}{\textbf{Naive-2-Shot}} 
    & \multicolumn{6}{c}{\textbf{Naive-CoT}} \\
    
    \cmidrule(lr){2-7} \cmidrule(lr){8-13} \cmidrule(lr){14-19} \cmidrule(lr){20-25}
    \textbf{Model} &\textbf{$S_a$} & \textbf{$S_c$}& \textbf{$S_m$} & \textbf{$S_f$}  & 
    \textbf{$S_n$}  &\textbf{$Avg$}
   &\textbf{$S_a$} & \textbf{$S_c$}& \textbf{$S_m$} & \textbf{$S_f$}  & 
    \textbf{$S_n$}  &\textbf{$Avg$}
    &\textbf{$S_a$} & \textbf{$S_c$}& \textbf{$S_m$} & \textbf{$S_f$}  & 
    \textbf{$S_n$}  &\textbf{$Avg$}
   &\textbf{$S_a$} & \textbf{$S_c$}& \textbf{$S_m$} & \textbf{$S_f$}  & 
    \textbf{$S_n$}  &\textbf{$Avg$} \\
     \midrule
     \multicolumn{8}{l}{\textbf{\textit{Chinese-oriented Models}}} \\
     \midrule
     ChatGLM3-6B& 0.56 & 0.20 & 0.72 & 0.73 & 0.71 & 0.58&
                0.59 & 0.13 & 0.72 & 0.68 & 0.67 & 0.56&
                0.62 & 0.13 & 0.72 & 0.68 & 0.68 & 0.57&
                0.64 & 0.16 & 0.71 & 0.69 & 0.67 & 0.57\\
     Qwen1.5-7B-Chat & 0.63 & 0.15 & 0.72 & 0.68 & 0.71 & 0.58 
     & 0.66 & 0.13 & 0.72 & 0.70 & \underline{0.71} & 0.59 & 0.67 & 0.13 & 0.71 & 0.70 & \underline{0.69} & 0.58 & 0.67 & 0.13 & 0.72 & 0.69 & 0.69 & 0.59\\
    Qwen1.5-14B-Chat & 0.66 & 0.16 & 0.72 & 0.69 & 0.73 & 0.60 & 0.68 & 0.17 & 0.72 & 0.67 & 0.71 & 0.60 & 0.74 & 0.18 & 0.71 & 0.71 & 0.65 & 0.60 & 0.69 & 0.18 & 0.72 & 0.73 & 0.68 & 0.60\\
    Qwen1.5-72B-Chat & 0.72 & 0.16 & 0.71 & 0.71 & 0.67 & 0.60 & 0.67 & 0.21 & 0.72 & 0.72 & 0.67 & 0.60 & 0.63 & 0.18 & 0.72 & 0.71 & 0.65 & 0.58 & \underline{0.78} & 0.20 & 0.70 & 0.71 & 0.65 & 0.61\\
    Deepseek-R1 & 0.70& \textbf{0.39}& 0.72& \textbf{0.76}& 0.49& \underline{0.61}&0.67& \textbf{0.38}& 0.71& 0.71& 0.54& \underline{0.60}&0.71& \textbf{0.38}& 0.71& 0.72& 0.54 &\underline{0.62}&0.77 &\textbf{0.35} &0.71 &\textbf{0.74}& 0.54 &\underline{0.62}\\

     \midrule
     \multicolumn{8}{l}{\textbf{\textit{English-oriented Models}}} \\
     \midrule
    Mixture-7B-v0.2 & \underline{0.77} & 0.08 & 0.70 & 0.74 & 0.55 & 0.57 & \textbf{0.82} & 0.17 & 0.71 & \textbf{0.75} & 0.52 & 0.59 & \textbf{0.82} & 0.15 & 0.70 & \underline{0.75} & 0.46 & 0.58 & 0.77 & 0.09 & 0.71 & 0.73 & 0.58 & 0.58\\
    Llama2-7B-Chat-hf &0.46 & 0.15 & \underline{0.73} & 0.73 & \underline{0.74} & 0.56 & 0.46 & 0.09 & \underline{0.73} & 0.71 & 0.66 & 0.53 & 0.44 & 0.12 & \textbf{0.73} & 0.74 & 0.67 & 0.54 & 0.49 & 0.14 & \textbf{0.74} & 0.73 & \underline{0.70} & 0.56\\
    Llama2-13B-Chat-hf &0.48 & 0.15 & \textbf{0.74} & 0.72 & \textbf{0.74} & 0.56 & 0.44 & 0.10 & \textbf{0.74} & 0.72 & \textbf{0.74} & 0.56 & 0.50 & 0.13 & \underline{0.73} & 0.68 & \textbf{0.74} & 0.57 & 0.45 & 0.10 & 0.73 & 0.67 & \textbf{0.74} & 0.55\\
    Llama2-70B-Chat-hf & 0.60 & 0.11 & 0.69 & 0.67 & 0.62 & 0.55 & 0.65 & 0.20 & 0.71 & 0.66 & 0.67 & 0.58 & 0.70 & 0.20 & 0.71 & 0.69 & 0.63 & 0.59 & 0.75 & 0.13 & 0.71 & 0.68 & 0.66 & 0.59\\
    \midrule
\multicolumn{8}{l}{\textbf{\textit{Close-source Models}}} \\
\midrule
    GPT-3.5-turbo & 0.62 & 0.11 & 0.71 & 0.72 & 0.62 & 0.56 & 0.72 & 0.16 & 0.71 & 0.75 & 0.59 & 0.59 & 0.73 & 0.14 & 0.71 & 0.74& 0.57 & 0.58& 0.76&0.10& 0.71 & 0.70 & 0.58 & 0.57\\
    GPT-4o & \textbf{0.79}& \underline{0.23} &0.71 &\underline{0.74} &0.58& \textbf{0.61} & \underline{0.75} &\underline{0.24} & 0.70 & \underline{0.74} & 0.60 & \textbf{0.61} & \underline{0.80} & \underline{0.23} & 0.71 & \textbf{0.76} & 0.57 & \textbf{0.62}& \textbf{0.83}&\underline{0.22}& \underline{0.71} & \underline{0.73} & 0.60 & \textbf{0.62}\\
    \bottomrule
\end{tabular}
    }
    \caption{
    Comparison of performance of various models on our evaluation system for QG tasks in in Naive-0-shot, Naive-1-shot, Naive-2-shot and Naive-cot settings.
    In these naive experimental setup, our experiment does not employ RAG or rerank metrics. Instead, it relies solely on a specifically designed prompt to enable the models to execute the QG task.
    The prompt for each setting is detailed in the Appendix ~\ref{naive}.
    }
    \label{tab:result3}
\end{table*}


\subsection{Results}
We conduct experiments on models of different ranges and sizes on our benchmark, and the results are shown in Tab.~\ref{tab:result1}. 
For more detailed analysis, please refer to Appendix ~\ref{result}.
\paragraph{Severity of Quotaiton Hallucination}

The results show that more than half of the citations generated by LLaMA2-13B-Chat are not genuine quotes. Furthermore, 
despite varying parameter sizes, all models demonstrate suboptimal performance on the QR task, especially on the $S_c$ metric. 
Even the best-performing model, GPT-4o, only scores 0.23 on the $S_c$ indicator, highlighting the critical need to address the quotation hallucination problem. 

\paragraph{Performance of Quotation-specific Reranking Metric}
Notably, our Quotation-specific Reranking method achieves the best results in each indicator, demonstrating the effectiveness of our method.
Since our method retrieves the most relevant and appropriate citations from the quotation database, it ensures the authenticity and credibility of the citations. Therefore, both $S_a$ and $S_c$ are equal to 1.
In addition, our method can effectively improve the novelty of citations and alleviate the problem of generating common citations with LLMs.

\paragraph*{Comparison between Model Sizes}
We conduct further analysis on different model sizes. Within the same series, larger models tend to show improved performance. This indicates that larger models have richer quotation memory and stronger instruction-following capabilities.

\subsection{Ablation Study}
\paragraph*{Correlations with Human Ratings} 
We randomly select five samples for each scenario from the evaluation dataset, totaling 105 data samples. Due to the varying requirements for background knowledge across different categories (Fig.\ref{fig:dataset}), this study specifically invites expert professors in relevant fields to manually score evaluation metrics. 
Since Quotation Authenticity and Credibility are objective factual metrics, the manual evaluation primarily focused on the remaining three metrics.
The process is independently conducted by experts, who are free to consult relevant literature and materials during the evaluation to ensure the reliability and objectivity of the results.
Subsequently, we employ correlation analysis to assess the degree of association between various metrics and the overall evaluation results. 
As shown in Fig.~\ref{fig:070relation}, all metrics exhibit high levels of correlation. Specifically, the correlation coefficients are significantly higher than the threshold for statistical significance, indicating that our metric system effectively reflects the actual conditions of the evaluation subjects. 
For correlation analyses of specific categories, please refer to the Appendix~\ref{sec:human}, where the results also reveal a significant correlation between manual and automated metrics for each category.

\paragraph*{Correlation between Evaluation Metrics} We present the correlations among the five automatic metrics in Tab.~\ref{tab:correlation_highlight}. As shown, the correlations between the metrics are all weak. This indicates that the five metrics are mutually independent, making it necessary to evaluate each of them individually in order to obtain a comprehensive view of the citation generation task assessment.
\newcolumntype{b}{>{\columncolor{blue!10}}r}
\newcolumntype{d}{>{\columncolor{brown!10}}r}
\newcolumntype{q}{>{\columncolor{Green!10}}r}

\setlength\tabcolsep{1pt}
\begin{table}[h]
    \centering
    \tiny
    \resizebox{0.32\textwidth}{!}{
    \begin{tabular}{lccccc}
    \toprule
    \textbf{Metric} & \textbf{$S_a$} & \textbf{$S_c$} & \textbf{$S_m$} & \textbf{$S_f$} & \textbf{$S_n$} \\
    \midrule
\textbf{$S_a$} & 1.000 & -0.038 & -0.018 & -0.132 & 0.077 \\
\textbf{$S_c$}  & -0.038 & 1.000 & -0.033 & 0.025 & 0.005 \\
\textbf{$S_m$}      & -0.018 & -0.033 & 1.000 & 0.070 & 0.004 \\
\textbf{$S_f$}       & -0.132 & 0.025 & 0.070 & 1.000 & 0.002 \\
\textbf{$S_n$}      & 0.077 & 0.005 & 0.004 & 0.002 & 1.000 \\
    \bottomrule
    \end{tabular}
    }
    \caption{Correlation Matrix between Evaluation Metrics}
    \label{tab:correlation_highlight}
\end{table}

\paragraph*{Effectiveness of Reranking Metrics} 


This study delves into the effectiveness of the rerank metric designed in our method and validates it through a series of ablation experiments. 
We adopt the following metrics: Hit Ratio at rank K (\textbf{HR@K}(K=1,3)), Normalized Discounted Cumulative Gain at rank K (\textbf{NDCG@K}(K=1,3)), and Mean Reciprocal Rank (\textbf{MRR}) for comparison.
On our benchmark, we compare a range of defined quotation-rerank metrics with state-of-the-art supervised, unsupervised, and closed-source API-based reranking methods. The supervised baselines include: BM25~\cite{rerankbert} and monoT5~\cite{monot5}. The unsupervised baselines comprise UPR~\cite{upr} and bge-reranker-large~\cite{bge_reranker_large}. The closed-source API-based baselines include ChatGPT3.5 and ChatGPT4. 
As shown in Table~\ref{tab:result2}, our simple yet effective quotation reranking metrics that demonstrate strong performance across various evaluation criteria. Notably, the $PPL_\text{avg} \text{+} \text{Novelty}$ metric excels among the four metrics and ranks just behind GPT-4 in the nDCG@3 metric.
Importantly, both supervised and unsupervised methods underperform compared to our proposed reranking metrics. This indicates that our approach effectively captures the nuances of the quotation generation task, leading to superior citation recommendations.

\paragraph*{Comparison between Prompt Strategies} 
We compare various prompting methods for QG tasks, including 0-shot, 1-shot, 2-shot, and Chain of Thought (CoT)~\cite{wei2023chainofthoughtpromptingelicitsreasoning} strategies. For the CoT method, we implement a basic "let's think step-by-step" approach. As shown in Table~\ref{tab:result3}, among the four naive settings, the CoT method outperforms the others . The performance variations among the few-shot settings are not statistically significant, which suggests that the model's in-context learning~\cite{dong2024surveyincontextlearning} capability will not substantially enhance its quotation performance. In contrast, the logical reasoning stimulated by the CoT method improves the model's quotation abilities to a certain degree.

\begin{table*}[!htbp]
\centering

\resizebox{\textwidth}{!}{
\begin{tabular}{|c|cc|}
\hline
  \multirow{1}{*}{\textbf{Method}} &
  \multicolumn{1}{c}{\multirow{1}{*}{\textbf{Literal Sentence}}} &
  \multirow{1}{*}
  {\begin{tabular}[c]{|p{5cm}|p{5cm}|p{5cm}}
  \textbf{Recalled List} &
  \textbf{Metric Rerank} &
  \textbf{Human Rerank}
  
  \end{tabular}}
  \\
  \hline
  \multirow{5}{*}{BM25}
  &
  \multirow{5}{*}{\begin{tabular}[c]{@{}p{10cm}@{}}
   Education empowers individuals to transform their lives and contribute to societal progress. [Q]. It fosters critical thinking, innovation, and social responsibility. By providing access to knowledge, education breaks down barriers and creates opportunities. It is a key driver of positive change and development. \end{tabular}} 
  &
  
  {\begin{tabular}[c]{|p{5cm}|p{5cm}|p{5cm}}
  \cellcolor{green3}Education is a human right with immense power to transform. On its foundation rest the cornerstones of freedom, democracy and sustainable human development.
  &
  \cellcolor{green4}Education is the transmission of civilization.
  &
  \cellcolor{green1}Education is the most powerful weapon which you can use to change the world
  \\
  \hline
  \cellcolor{green4}Education is the transmission of civilization &
  \cellcolor{green3}Education is a human right with immense power to transform. On its foundation rest the cornerstones of freedom, democracy and sustainable human development &
  \cellcolor{green2}Knowledge is power. Information is liberating. Education is the premise of progress, in every society, in every family.
   \\
   \hline
  \cellcolor{green2}Knowledge is power. Information is liberating. Education is the premise of progress, in every society, in every family &
  \cellcolor{green1}Education is the most powerful weapon which you can use to change the world  &
   \cellcolor{green3}Education is a human right with immense power to transform. On its foundation rest the cornerstones of freedom, democracy and sustainable human development
   \\
    \hline
   \cellcolor{green1}Education is the most powerful weapon which you can use to change the world  &
  \cellcolor{green2}Knowledge is power. Information is liberating. Education is the premise of progress, in every society, in every family &
   \cellcolor{green4}Education is the transmission of civilization
   \\
    \hline
   \cellcolor{green5}The function of education is to teach one to think intensively and to think critically. Intelligence plus character - that is the goal of true education &
  \cellcolor{green5}The function of education is to teach one to think intensively and to think critically. Intelligence plus character - that is the goal of true education &
  \cellcolor{green5}The function of education is to teach one to think intensively and to think critically. Intelligence plus character - that is the goal of true education 
  \end{tabular}}
  \\
  \hline
  \multirow{5}{*}{Ours} &
  \multirow{5}{*}{\begin{tabular}[c]{@{}p{10cm}@{}}
  Education empowers individuals to transform their lives and contribute to societal progress. [Q]. It fosters critical thinking, innovation, and social responsibility. By providing access to knowledge, education breaks down barriers and creates opportunities. It is a key driver of positive change and development.
  \end{tabular}}
  &
  {\begin{tabular}[c]{|p{5cm}|p{5cm}|p{5cm}}
  \cellcolor{green3}Education is a human right with immense power to transform. On its foundation rest the cornerstones of freedom, democracy and sustainable human development &
  \cellcolor{green1}Education is the most powerful weapon which you can use to change the world &
  \cellcolor{green1}Education is the most powerful weapon which you can use to change the world
  \\
  \hline
  \cellcolor{green4}Education is the transmission of civilization &
  \cellcolor{green3}Education is a human right with immense power to transform. On its foundation rest the cornerstones of freedom, democracy and sustainable human development &
  \cellcolor{green2}Knowledge is power. Information is liberating. Education is the premise of progress, in every society, in every family
  \\
  \hline
  \cellcolor{green2}Knowledge is power. Information is liberating. Education is the premise of progress, in every society, in every family &
  \cellcolor{green2}Knowledge is power. Information is liberating. Education is the premise of progress, in every society, in every family &
  \cellcolor{green3}Education is a human right with immense power to transform. On its foundation rest the cornerstones of freedom, democracy and sustainable human development
  \\
  \hline
  \cellcolor{green1} Education is the most powerful weapon which you can use to change the world &
  \cellcolor{green4}Education is the transmission of civilization &
  \cellcolor{green4}Education is the transmission of civilization 
  \\
  \hline
  \cellcolor{green5} The function of education is to teach one to think intensively and to think critically. Intelligence plus character - that is the goal of true education &
  \cellcolor{green5} The function of education is to teach one to think intensively and to think critically. Intelligence plus character - that is the goal of true education &
 \cellcolor{green5} The function of education is to teach one to think intensively and to think critically. Intelligence plus character - that is the goal of true education
\end{tabular}}
  \\
  \hline

\end{tabular}
}
\caption{ The examples of recalled candidates reranked via different rerank metrics and human evaluation. 
The indicators [Q] denotes the insertion positions of the given context. A darker shade of green indicates a higher rank bestowed by humans. See the Appendix for a detailed
comparison of the unsupervised UPR, the closed-
source model GPT-3.5-turbo, and our approach.
}
\label{tab:application}
\end{table*}

\subsection{QUILL Application}
In this study, we conduct a comprehensive case analysis to demonstrate the efficacy and alignment of our reranking metric with human evaluations. As illustrated in Tab.~\ref{tab:application}, we focus on several key models for comparison: the supervised BM25 and our own reranking metric, which combines average perplexity ($\text{PPL}_{avg}$) with novelty. Additionally, we manually sort and annotate the top-5 quote list initially recalled, serving as a benchmark for comparison. The findings reveal that our metric exhibits a higher correlation with human sorting than the other methods, underscoring its broad applicability and effectiveness. See the Appendix for a detailed comparison of the unsupervised UPR, the closed-source model GPT-3.5-turbo, and our approach.


\section{Conclusion}


In this paper, we systematically explore methods to enhance the performance of quotation generation tasks in LLMs.
Initially, we establish a holistic and automatic evaluation system consisting of five highly interpretable and rigorous criteria , facilitating both human and automatic evaluation of this task.
Then, we construct a comprehensive and high-quality knowledge database containing up to 32,022 quotes, complete with authors or sources. 
Moreover, we design a fine-grained quotation-specific metric to rerank the retrieved quotations from the knowledge base to improve QG performance.
Additionally, we conduct extensive experiments to verify that our metrics are strongly correlate with human preference and significantly effective in both open-source and closed-source LLMs.

\section*{Limitations}
This study highlights several limitations. We primarily use Perplexity (PPL) to evaluate text fluency. Although PPL is widely applied, it only measures the divergence between the model's and true probability distributions. Future research should integrate additional metrics or human evaluations for a more comprehensive assessment.
Additionally, our analysis is restricted to specific contexts with clear correlations before and after quoted content. While informative, this approach does not cover a wide range of quoting scenarios. Future studies should explore diverse applications for more generalizable insights.

\bibliography{custom}

\begin{thebibliography}{44}
\providecommand{\natexlab}[1]{#1}

\bibitem[{Achiam et~al.(2023)Achiam, Adler, Agarwal, Ahmad, Akkaya, Aleman, Almeida, Altenschmidt, Altman, Anadkat et~al.}]{achiam2023gpt}
Josh Achiam, Steven Adler, Sandhini Agarwal, Lama Ahmad, Ilge Akkaya, Florencia~Leoni Aleman, Diogo Almeida, Janko Altenschmidt, Sam Altman, Shyamal Anadkat, et~al. 2023.
\newblock Gpt-4 technical report.
\newblock \emph{arXiv preprint arXiv:2303.08774}.

\bibitem[{Agrawal et~al.(2024)Agrawal, Suzgun, Mackey, and Kalai}]{agrawal2024languagemodelsknowtheyre}
Ayush Agrawal, Mirac Suzgun, Lester Mackey, and Adam~Tauman Kalai. 2024.
\newblock \href {https://arxiv.org/abs/2305.18248} {Do language models know when they're hallucinating references?}
\newblock \emph{Preprint}, arXiv:2305.18248.

\bibitem[{Ahn et~al.(2016)Ahn, Lee, Jeon, Ha, and goo Lee}]{Ahn2016}
Yeonchan Ahn, Hanbit Lee, Heesik Jeon, Seungdo Ha, and Sang goo Lee. 2016.
\newblock \href {https://api.semanticscholar.org/CorpusID:17252129} {Quote recommendation for dialogs and writings}.
\newblock In \emph{CBRecSys@RecSys}.

\bibitem[{Anil et~al.(2023)Anil, Dai, Firat, Johnson, Lepikhin, Passos, Shakeri, Taropa, Bailey, Chen et~al.}]{anil2023palm}
Rohan Anil, Andrew~M Dai, Orhan Firat, Melvin Johnson, Dmitry Lepikhin, Alexandre Passos, Siamak Shakeri, Emanuel Taropa, Paige Bailey, Zhifeng Chen, et~al. 2023.
\newblock Palm 2 technical report.
\newblock \emph{arXiv preprint arXiv:2305.10403}.

\bibitem[{BAAI(2023)}]{bge_reranker_large}
BAAI. 2023.
\newblock Bge-reranker-large: A pre-trained model for ranking tasks.
\newblock \url{https://huggingface.co/BAAI/bge-reranker-large}.

\bibitem[{Bai et~al.(2023)Bai, Bai, Chu, Cui, Dang, Deng, Fan, Ge, Han, Huang et~al.}]{bai2023qwen}
Jinze Bai, Shuai Bai, Yunfei Chu, Zeyu Cui, Kai Dang, Xiaodong Deng, Yang Fan, Wenbin Ge, Yu~Han, Fei Huang, et~al. 2023.
\newblock Qwen technical report.
\newblock \emph{arXiv preprint arXiv:2309.16609}.

\bibitem[{Bang et~al.(2023)Bang, Cahyawijaya, Lee, Dai, Su, Wilie, Lovenia, Ji, Yu, Chung, Do, Xu, and Fung}]{bang2023multitaskmultilingualmultimodalevaluation}
Yejin Bang, Samuel Cahyawijaya, Nayeon Lee, Wenliang Dai, Dan Su, Bryan Wilie, Holy Lovenia, Ziwei Ji, Tiezheng Yu, Willy Chung, Quyet~V. Do, Yan Xu, and Pascale Fung. 2023.
\newblock \href {https://arxiv.org/abs/2302.04023} {A multitask, multilingual, multimodal evaluation of chatgpt on reasoning, hallucination, and interactivity}.
\newblock \emph{Preprint}, arXiv:2302.04023.

\bibitem[{Chen et~al.(2023)Chen, Pasupat, Singh, Lee, and Guu}]{chen2023purrefficientlyeditinglanguage}
Anthony Chen, Panupong Pasupat, Sameer Singh, Hongrae Lee, and Kelvin Guu. 2023.
\newblock \href {https://arxiv.org/abs/2305.14908} {Purr: Efficiently editing language model hallucinations by denoising language model corruptions}.
\newblock \emph{Preprint}, arXiv:2305.14908.

\bibitem[{Chern et~al.(2023)Chern, Chern, Chen, Yuan, Feng, Zhou, He, Neubig, and Liu}]{chern2023factoolfactualitydetectiongenerative}
I-Chun Chern, Steffi Chern, Shiqi Chen, Weizhe Yuan, Kehua Feng, Chunting Zhou, Junxian He, Graham Neubig, and Pengfei Liu. 2023.
\newblock \href {https://arxiv.org/abs/2307.13528} {Factool: Factuality detection in generative ai -- a tool augmented framework for multi-task and multi-domain scenarios}.
\newblock \emph{Preprint}, arXiv:2307.13528.

\bibitem[{Cho et~al.(2014)Cho, van Merrienboer, Gulcehre, Bahdanau, Bougares, Schwenk, and Bengio}]{cho2014learningphraserepresentationsusing}
Kyunghyun Cho, Bart van Merrienboer, Caglar Gulcehre, Dzmitry Bahdanau, Fethi Bougares, Holger Schwenk, and Yoshua Bengio. 2014.
\newblock \href {https://arxiv.org/abs/1406.1078} {Learning phrase representations using rnn encoder-decoder for statistical machine translation}.
\newblock \emph{Preprint}, arXiv:1406.1078.

\bibitem[{Dale et~al.(2022)Dale, Voita, Barrault, and Costa-jussà}]{dale2022detectingmitigatinghallucinationsmachine}
David Dale, Elena Voita, Loïc Barrault, and Marta~R. Costa-jussà. 2022.
\newblock \href {https://arxiv.org/abs/2212.08597} {Detecting and mitigating hallucinations in machine translation: Model internal workings alone do well, sentence similarity even better}.
\newblock \emph{Preprint}, arXiv:2212.08597.

\bibitem[{Devlin et~al.(2019)Devlin, Chang, Lee, and Toutanova}]{devlin2019bertpretrainingdeepbidirectional}
Jacob Devlin, Ming-Wei Chang, Kenton Lee, and Kristina Toutanova. 2019.
\newblock \href {https://arxiv.org/abs/1810.04805} {Bert: Pre-training of deep bidirectional transformers for language understanding}.
\newblock \emph{Preprint}, arXiv:1810.04805.

\bibitem[{Dong et~al.(2024)Dong, Li, Dai, Zheng, Ma, Li, Xia, Xu, Wu, Chang, Sun, Li, and Sui}]{dong2024surveyincontextlearning}
Qingxiu Dong, Lei Li, Damai Dai, Ce~Zheng, Jingyuan Ma, Rui Li, Heming Xia, Jingjing Xu, Zhiyong Wu, Baobao Chang, Xu~Sun, Lei Li, and Zhifang Sui. 2024.
\newblock \href {https://arxiv.org/abs/2301.00234} {A survey on in-context learning}.
\newblock \emph{Preprint}, arXiv:2301.00234.

\bibitem[{Filippova(2020)}]{filippova-2020-controlled}
Katja Filippova. 2020.
\newblock \href {https://doi.org/10.18653/v1/2020.findings-emnlp.76} {Controlled hallucinations: Learning to generate faithfully from noisy data}.
\newblock In \emph{Findings of the Association for Computational Linguistics: EMNLP 2020}, pages 864--870, Online. Association for Computational Linguistics.

\bibitem[{Guerreiro et~al.(2023)Guerreiro, Alves, Waldendorf, Haddow, Birch, Colombo, and Martins}]{guerreiro2023hallucinationslargemultilingualtranslation}
Nuno~M. Guerreiro, Duarte Alves, Jonas Waldendorf, Barry Haddow, Alexandra Birch, Pierre Colombo, and André F.~T. Martins. 2023.
\newblock \href {https://arxiv.org/abs/2303.16104} {Hallucinations in large multilingual translation models}.
\newblock \emph{Preprint}, arXiv:2303.16104.

\bibitem[{Han et~al.(2024)Han, Yang, Peng, Tiwari, Wan, Liu, and Wang}]{han2024empiricalstudyinformationextraction}
Ridong Han, Chaohao Yang, Tao Peng, Prayag Tiwari, Xiang Wan, Lu~Liu, and Benyou Wang. 2024.
\newblock \href {https://arxiv.org/abs/2409.00369} {An empirical study on information extraction using large language models}.
\newblock \emph{Preprint}, arXiv:2409.00369.

\bibitem[{Huang et~al.(2023)Huang, Yu, Ma, Zhong, Feng, Wang, Chen, Peng, Feng, Qin, and Liu}]{huang2023surveyhallucinationlargelanguage}
Lei Huang, Weijiang Yu, Weitao Ma, Weihong Zhong, Zhangyin Feng, Haotian Wang, Qianglong Chen, Weihua Peng, Xiaocheng Feng, Bing Qin, and Ting Liu. 2023.
\newblock \href {https://arxiv.org/abs/2311.05232} {A survey on hallucination in large language models: Principles, taxonomy, challenges, and open questions}.
\newblock \emph{Preprint}, arXiv:2311.05232.

\bibitem[{Karpukhin et~al.(2020)Karpukhin, Oguz, Min, Lewis, Wu, Edunov, Chen, and Yih}]{karpukhin-etal-2020-dense}
Vladimir Karpukhin, Barlas Oguz, Sewon Min, Patrick Lewis, Ledell Wu, Sergey Edunov, Danqi Chen, and Wen-tau Yih. 2020.
\newblock \href {https://doi.org/10.18653/v1/2020.emnlp-main.550} {Dense passage retrieval for open-domain question answering}.
\newblock In \emph{Proceedings of the 2020 Conference on Empirical Methods in Natural Language Processing (EMNLP)}, pages 6769--6781, Online. Association for Computational Linguistics.

\bibitem[{Kington et~al.(2021)Kington, Arnesen, Chou, Curry, Lazer, and Villarruel}]{kington2021identifying}
Raynard~S Kington, Stacey Arnesen, Wen-Ying~Sylvia Chou, Susan~J Curry, David Lazer, and Antonia~M Villarruel. 2021.
\newblock Identifying credible sources of health information in social media: principles and attributes.
\newblock \emph{NAM perspectives}, 2021.

\bibitem[{Krumm et~al.(2020)Krumm, Berres, Kivisaari, Monsch, Reinhardt, Blatow, Kressig, and Taylor}]{10.1093/arclin/acaa109}
Sabine Krumm, Manfred Berres, Sasa~L Kivisaari, Andreas~U Monsch, Julia Reinhardt, Maria Blatow, Reto~W Kressig, and Kirsten~I Taylor. 2020.
\newblock \href {https://doi.org/10.1093/arclin/acaa109} {Cats and apples: Semantic fluency performance for living things identifies patients with very early alzheimer’s disease}.
\newblock \emph{Archives of Clinical Neuropsychology}, 36(5):838--843.

\bibitem[{Lee et~al.(2016)Lee, Ahn, Lee, Ha, and Lee}]{10.1145/2911451.2914734}
Hanbit Lee, Yeonchan Ahn, Haejun Lee, Seungdo Ha, and Sang-goo Lee. 2016.
\newblock \href {https://doi.org/10.1145/2911451.2914734} {Quote recommendation in dialogue using deep neural network}.
\newblock In \emph{Proceedings of the 39th International ACM SIGIR Conference on Research and Development in Information Retrieval}, SIGIR '16, page 957–960, New York, NY, USA. Association for Computing Machinery.

\bibitem[{Lewis et~al.(2021)Lewis, Perez, Piktus, Petroni, Karpukhin, Goyal, Küttler, Lewis, tau Yih, Rocktäschel, Riedel, and Kiela}]{lewis2021retrievalaugmentedgenerationknowledgeintensivenlp}
Patrick Lewis, Ethan Perez, Aleksandra Piktus, Fabio Petroni, Vladimir Karpukhin, Naman Goyal, Heinrich Küttler, Mike Lewis, Wen tau Yih, Tim Rocktäschel, Sebastian Riedel, and Douwe Kiela. 2021.
\newblock \href {https://arxiv.org/abs/2005.11401} {Retrieval-augmented generation for knowledge-intensive nlp tasks}.
\newblock \emph{Preprint}, arXiv:2005.11401.

\bibitem[{MacLaughlin and Smith(2021)}]{maclaughlin-smith-2021-content}
Ansel MacLaughlin and David Smith. 2021.
\newblock \href {https://doi.org/10.18653/v1/2021.eacl-main.195} {Content-based models of quotation}.
\newblock In \emph{Proceedings of the 16th Conference of the European Chapter of the Association for Computational Linguistics: Main Volume}, pages 2296--2314, Online. Association for Computational Linguistics.

\bibitem[{Manakul et~al.(2023)Manakul, Liusie, and Gales}]{manakul2023selfcheckgptzeroresourceblackboxhallucination}
Potsawee Manakul, Adian Liusie, and Mark J.~F. Gales. 2023.
\newblock \href {https://arxiv.org/abs/2303.08896} {Selfcheckgpt: Zero-resource black-box hallucination detection for generative large language models}.
\newblock \emph{Preprint}, arXiv:2303.08896.

\bibitem[{Maynez et~al.(2020)Maynez, Narayan, Bohnet, and McDonald}]{maynez-etal-2020-faithfulness}
Joshua Maynez, Shashi Narayan, Bernd Bohnet, and Ryan McDonald. 2020.
\newblock \href {https://doi.org/10.18653/v1/2020.acl-main.173} {On faithfulness and factuality in abstractive summarization}.
\newblock In \emph{Proceedings of the 58th Annual Meeting of the Association for Computational Linguistics}, pages 1906--1919, Online. Association for Computational Linguistics.

\bibitem[{Mündler et~al.(2024)Mündler, He, Jenko, and Vechev}]{mündler2024selfcontradictoryhallucinationslargelanguage}
Niels Mündler, Jingxuan He, Slobodan Jenko, and Martin Vechev. 2024.
\newblock \href {https://arxiv.org/abs/2305.15852} {Self-contradictory hallucinations of large language models: Evaluation, detection and mitigation}.
\newblock \emph{Preprint}, arXiv:2305.15852.

\bibitem[{Nogueira et~al.(2020)Nogueira, Jiang, Pradeep, and Lin}]{monot5}
Rodrigo Nogueira, Zhiying Jiang, Ronak Pradeep, and Jimmy Lin. 2020.
\newblock \href {https://doi.org/10.18653/v1/2020.findings-emnlp.63} {Document ranking with a pretrained sequence-to-sequence model}.
\newblock In \emph{Findings of the Association for Computational Linguistics: EMNLP 2020}, pages 708--718, Online. Association for Computational Linguistics.

\bibitem[{Nogueira and Cho(2019)}]{rerankbert}
Rodrigo~Frassetto Nogueira and Kyunghyun Cho. 2019.
\newblock \href {https://arxiv.org/abs/1901.04085} {Passage re-ranking with {BERT}}.
\newblock \emph{CoRR}, abs/1901.04085.

\bibitem[{OpenAI(2022)}]{chatgpt}
OpenAI. 2022.
\newblock \href {https://openai.com/index/chatgpt/} {Introducing chatgpt}.

\bibitem[{Pfeiffer et~al.(2023)Pfeiffer, Piccinno, Nicosia, Wang, Reid, and Ruder}]{pfeiffer2023mmt5modularmultilingualpretraining}
Jonas Pfeiffer, Francesco Piccinno, Massimo Nicosia, Xinyi Wang, Machel Reid, and Sebastian Ruder. 2023.
\newblock \href {https://arxiv.org/abs/2305.14224} {mmt5: Modular multilingual pre-training solves source language hallucinations}.
\newblock \emph{Preprint}, arXiv:2305.14224.

\bibitem[{Qi et~al.(2022{\natexlab{a}})Qi, Yang, Yi, Cheng, Liu, and Sun}]{qi-etal-2022-quoter}
Fanchao Qi, Yanhui Yang, Jing Yi, Zhili Cheng, Zhiyuan Liu, and Maosong Sun. 2022{\natexlab{a}}.
\newblock \href {https://doi.org/10.18653/v1/2022.acl-long.27} {{Q}uote{R}: A benchmark of quote recommendation for writing}.
\newblock In \emph{Proceedings of the 60th Annual Meeting of the Association for Computational Linguistics (Volume 1: Long Papers)}, pages 336--348, Dublin, Ireland. Association for Computational Linguistics.

\bibitem[{Qi et~al.(2022{\natexlab{b}})Qi, Yang, Yi, Cheng, Liu, and Sun}]{qi2022}
Fanchao Qi, Yanhui Yang, Jing Yi, Zhili Cheng, Zhiyuan Liu, and Maosong Sun. 2022{\natexlab{b}}.
\newblock \href {https://aclanthology.org/2022.acl-long.27} {Quoter: A benchmark of quote recommendation for writing}.
\newblock In \emph{Proceedings of the 60th Annual Meeting of the Association for Computational Linguistics (Volume 1: Long Papers)}, pages 336--348.

\bibitem[{Quora(2020)}]{sm}
Quora. 2020.
\newblock \href {https://www.quora.com/What-happens-if-you-make-too-many-citation-mistakes-in-your-research-paper?__cf_chl_tk=DOWYnkh.2RbmLEerjtMDgr2J9CyZrgMt5BpxKY08y6g-1723737569-0.0.1.1-4116} {What happens if you make too many citation mistakes in your research paper?}

\bibitem[{Sachan et~al.(2023)Sachan, Lewis, Joshi, Aghajanyan, tau Yih, Pineau, and Zettlemoyer}]{upr}
Devendra~Singh Sachan, Mike Lewis, Mandar Joshi, Armen Aghajanyan, Wen tau Yih, Joelle Pineau, and Luke Zettlemoyer. 2023.
\newblock \href {https://arxiv.org/abs/2204.07496} {Improving passage retrieval with zero-shot question generation}.
\newblock \emph{Preprint}, arXiv:2204.07496.

\bibitem[{Savov(2021)}]{savov2021measuring}
Pavel Savov. 2021.
\newblock Measuring the novelty of scientific papers.

\bibitem[{Sridhar and Visser(2023)}]{sridhar2023improvedbeamsearchhallucination}
Arvind~Krishna Sridhar and Erik Visser. 2023.
\newblock \href {https://arxiv.org/abs/2212.02712} {Improved beam search for hallucination mitigation in abstractive summarization}.
\newblock \emph{Preprint}, arXiv:2212.02712.

\bibitem[{Tan et~al.(2015{\natexlab{a}})Tan, Wan, and Xiao}]{Tan_Wan_Xiao_2015}
Jiwei Tan, Xiaojun Wan, and Jianguo Xiao. 2015{\natexlab{a}}.
\newblock \href {https://doi.org/10.1609/aaai.v29i1.9530} {Learning to recommend quotes for writing}.
\newblock \emph{Proceedings of the AAAI Conference on Artificial Intelligence}, 29(1).

\bibitem[{Tan et~al.(2015{\natexlab{b}})Tan, Wan, and Xiao}]{Tan2005}
Jiwei Tan, Xiaojun Wan, and Jianguo Xiao. 2015{\natexlab{b}}.
\newblock \href {https://doi.org/10.1609/aaai.v29i1.9530} {Learning to recommend quotes for writing}.
\newblock \emph{Proceedings of the AAAI Conference on Artificial Intelligence}, 29(1).
\newblock [Online; accessed 2024-10-22].

\bibitem[{Touvron et~al.(2023)Touvron, Martin, Stone, Albert, Almahairi, Babaei, Bashlykov, Batra, Bhargava, Bhosale et~al.}]{touvron2023llama}
Hugo Touvron, Louis Martin, Kevin Stone, Peter Albert, Amjad Almahairi, Yasmine Babaei, Nikolay Bashlykov, Soumya Batra, Prajjwal Bhargava, Shruti Bhosale, et~al. 2023.
\newblock Llama 2: Open foundation and fine-tuned chat models.
\newblock \emph{arXiv preprint arXiv:2307.09288}.

\bibitem[{Vaswani et~al.(2023)Vaswani, Shazeer, Parmar, Uszkoreit, Jones, Gomez, Kaiser, and Polosukhin}]{vaswani2023attentionneed}
Ashish Vaswani, Noam Shazeer, Niki Parmar, Jakob Uszkoreit, Llion Jones, Aidan~N. Gomez, Lukasz Kaiser, and Illia Polosukhin. 2023.
\newblock \href {https://arxiv.org/abs/1706.03762} {Attention is all you need}.
\newblock \emph{Preprint}, arXiv:1706.03762.

\bibitem[{Wang et~al.(2020)Wang, Li, Zeng, Zhang, and Wong}]{wang-etal-2020-continuity}
Lingzhi Wang, Jing Li, Xingshan Zeng, Haisong Zhang, and Kam-Fai Wong. 2020.
\newblock \href {https://doi.org/10.18653/v1/2020.emnlp-main.538} {Continuity of topic, interaction, and query: Learning to quote in online conversations}.
\newblock In \emph{Proceedings of the 2020 Conference on Empirical Methods in Natural Language Processing (EMNLP)}, pages 6640--6650, Online. Association for Computational Linguistics.

\bibitem[{Wang et~al.(2021)Wang, Zeng, and Wong}]{wang-etal-2021-quotation}
Lingzhi Wang, Xingshan Zeng, and Kam-Fai Wong. 2021.
\newblock \href {https://doi.org/10.18653/v1/2021.acl-short.95} {Quotation recommendation and interpretation based on transformation from queries to quotations}.
\newblock In \emph{Proceedings of the 59th Annual Meeting of the Association for Computational Linguistics and the 11th International Joint Conference on Natural Language Processing (Volume 2: Short Papers)}, pages 754--758, Online. Association for Computational Linguistics.

\bibitem[{Wei et~al.(2023)Wei, Wang, Schuurmans, Bosma, Ichter, Xia, Chi, Le, and Zhou}]{wei2023chainofthoughtpromptingelicitsreasoning}
Jason Wei, Xuezhi Wang, Dale Schuurmans, Maarten Bosma, Brian Ichter, Fei Xia, Ed~Chi, Quoc Le, and Denny Zhou. 2023.
\newblock \href {https://arxiv.org/abs/2201.11903} {Chain-of-thought prompting elicits reasoning in large language models}.
\newblock \emph{Preprint}, arXiv:2201.11903.

\bibitem[{Zhang et~al.(2024)Zhang, Pan, Zhao, and Wang}]{zhang2024knowledgealignmentproblembridging}
Shuo Zhang, Liangming Pan, Junzhou Zhao, and William~Yang Wang. 2024.
\newblock \href {https://arxiv.org/abs/2305.13669} {The knowledge alignment problem: Bridging human and external knowledge for large language models}.
\newblock \emph{Preprint}, arXiv:2305.13669.

\end{thebibliography}
\clearpage

\appendix
\renewcommand{\thesection}{\Alph{section}} 
\section*{Appendix}
\section{Details of Evaluation Dataset}
We also conducted manual analysis on the Evaluation Dataset, selecting 275 quotes from numerous context-quote pairs, dividing into Chinese and English, which categories and scenarios details are shown in Figure.\ref{fig:dataset}. After statistics, there are 204 Chinese samples and 71 English samples, with a total of 144 Chinese and English authors.

\section{Details of Quotation Knowledge Base}



\begin{figure}[t] 
    \centering
        \includegraphics[width=0.48\textwidth]{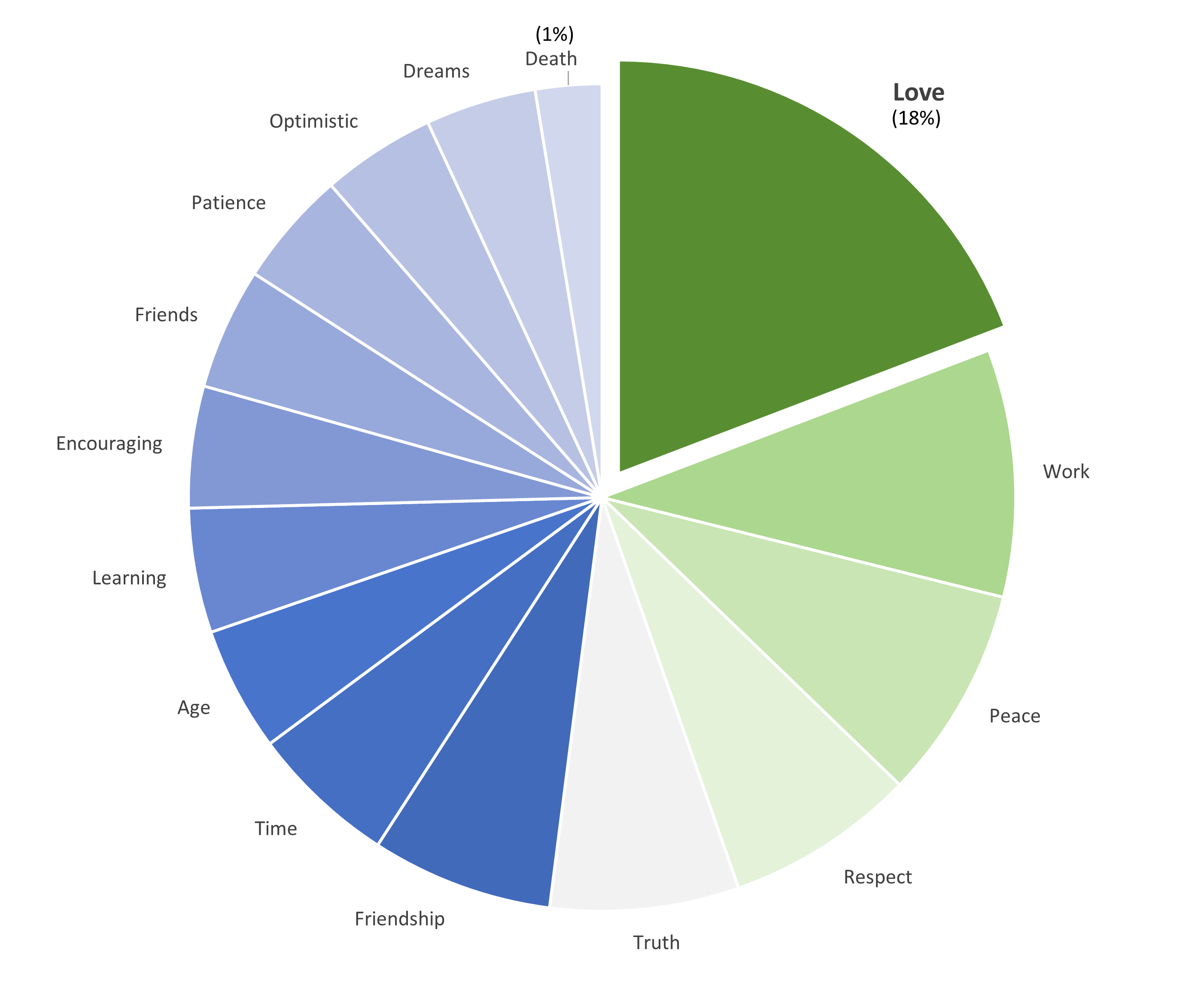}
    \caption{The specific topic distribution of the English quotation corpus.}
    \label{fig:060appendix1}
\end{figure}

\begin{figure}[t] 
    \centering
        \includegraphics[width=0.48\textwidth]{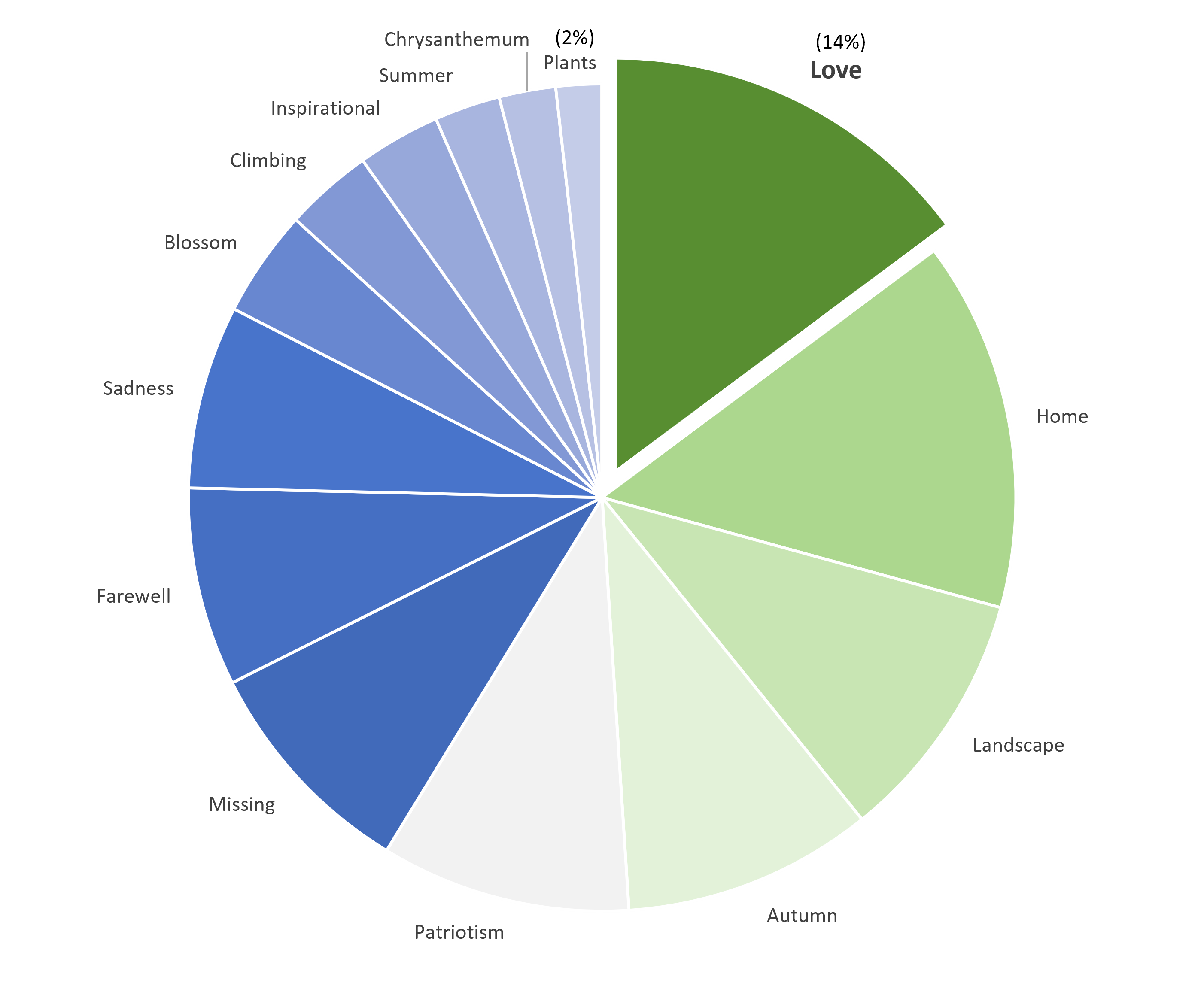}
    \caption{The specific topic distribution of the Classic Chinese quotation corpus.}
    \label{fig:060appendix3}
\end{figure}

This chapter further analyzes the data details in the quotation corpus, which is divided into three languages: English, Standard Chinese, and Classical Chinese, all classified by topic and author. The number of topics and authors for each language is shown in Table.\ref{tab:appendix1}.
In addition, we also conduct analysis on the proportion of different topics in each language in the corpus, as shown in Figure.
\ref{fig:060appendix1} -~\ref{fig:060appendix3}. for specific topics and proportions.

\newcolumntype{b}{>{\columncolor{blue!10}}r}
\newcolumntype{d}{>{\columncolor{brown!10}}r}
\newcolumntype{q}{>{\columncolor{Green!10}}r}

\setlength\tabcolsep{1pt}
\begin{table}[h]
    \centering
    \resizebox{0.38\textwidth}{!}{
    \begin{tabular}{lccb}
    \toprule
   \textbf{Language Type} & \textbf{Topic} & \textbf{Author} & \textbf{Total} \\
   \midrule
   English & 1,216 & 6,624 & 16,393\\
   Standard Chinese& 228 & 2,377 & 7,519\\
   Classic Chinese& 869 & 876 &  8,110\\
   \bottomrule 
    \end{tabular}
    }
    \caption{The specific topics, authors, and total count of the quotation corpus.}
    \label{tab:appendix1}
\end{table}

%
\section{Effectiveness of GPT-4o Extraction}
\label{sec:GPT-4o}

Previous studies have demonstrated that GPT-4o exhibits superior performance in simple information extraction tasks under zero-shot settings~\cite{han2024empiricalstudyinformationextraction}. In our context, we primarily extract authors and sources, which involves only two fields, thus categorizing this as a simple information extraction task. Therefore, we believe that GPT-4o is capable of achieving excellent extraction performance.
We also conduct experiments to validate the extraction effectiveness of GPT-4o in our task. We extract 100 citations containing authors or sources from a citation knowledge base in three different languages and use these as keywords to search on the Bing search engine. In the returned search results, GPT-4o is utilized to extract the authors or sources of the citations. Ultimately, we assess the matching degree between the fields extracted by GPT-4o and the annotated fields in the knowledge base. The specific results are as follows:
\newcolumntype{b}{>{\columncolor{blue!10}}r}
\newcolumntype{d}{>{\columncolor{brown!10}}r}
\newcolumntype{q}{>{\columncolor{Green!10}}r}

\setlength\tabcolsep{1pt}
\begin{table}[h]
    \centering
    \resizebox{0.35\textwidth}{!}{
    \begin{tabular}{lc}
    \toprule
   \textbf{Language Type} & \textbf{Extraction accuracy} \\
   \midrule
   English & 97\% \\
   Standard Chinese& 95\%\\
   Classic Chinese& 98\%\\
   \bottomrule 
    \end{tabular}
    }
    \caption{Extraction and verification results of ChatGPT}
    \label{tab:chatgpt}
\end{table}

\section{Details of the Inverted Sigmoid Function}
\label{sigmoid}


To map the calculated Perplexity (PPL) values to a range of [0, 1], this study employs the Sigmoid function, which not only maps the scores to [0, 1] but also handles positive extreme values in the data. For the two key parameters of the Sigmoid function, \(k\) and \(\mu\), the calculation methods used in this study are as follows:

For \(\mu\): The Sigmoid function outputs 0.5 when \(x = \mu\), where the slope is at its maximum. Typically, \(\mu\) is set to the median or mean of the data, ensuring that the middle values are mapped to 0.5. In this study, we choose the mean of the data as the value for \(\mu\).

For \(k\): The slope parameter \(k\) controls the "compression degree" of the mapping. A larger \(k\) results in a steeper Sigmoid curve, which is suitable for data with a concentrated distribution. In contrast, a smaller \(k\) results in a gentler curve, making it more appropriate for data with a wide range or extreme outliers. This study calculates \(k\) based on an empirical formula as follows:

\begin{equation}
\scriptsize
k= \dfrac{ln(9)}{Q_{95}-Q_5}
\end{equation}
where $ln(9) \approx$ 2.2, corresponding to the span of the Sigmoid function from 0.1 to 0.9, $Q_5$ represents the 5\% digit of the data, and $Q_{95}$ represents the 95\% digit of the data.

\section{More Analysis of Experimental Results}
\label{result}
Due to space limitations, we provide more experimental results analysis in this section.
\paragraph*{Comparison between Model Types}
The performance comparison between the Chinese-oriented group and the English-oriented group on the Chinese-English benchmark reveals no significant differences, suggesting that the model's quotation ability is not language-dependent.
Overall, the current opensource small to large-scale models exhibit a relatively small performance gap compared to close-source models, indicating the universality of the issue of quotation hallucination in LLMs.

\section{Details of Human Evaluation Metrics}
\label{sec:human}

We randomly selected 5 samples for each scenario from the evaluation dataset, totaling 105 data. Since different scenarios have different requirements for background knowledge, this study specially invited professional professors in related fields to manually score different categories of data. The scoring process was completed independently by experts, and relevant literature and materials were freely available during the review process to ensure the reliability and objectivity of the scoring results.
In addition, we also further analyzed the correlation analysis of each specific category. The results are shown in the Table.~\ref{tab:metric_results}. It can also be seen from the results that the manual indicators and automatic indicators of each category are also significantly correlated.

\newcolumntype{b}{>{\columncolor{blue!10}}r}
\newcolumntype{d}{>{\columncolor{brown!10}}r}
\newcolumntype{q}{>{\columncolor{Green!10}}r}

\setlength\tabcolsep{1pt}
\begin{table}[h]
    \centering
    \resizebox{0.5\textwidth}{!}{
    \begin{tabular}{lcccccccc}
    \toprule
    \textbf{Metric} & \textbf{Overall} & \textbf{Science} & \textbf{Business} & \textbf{News} & \textbf{Academic} & \textbf{Life} & \textbf{Law} & \textbf{Diplomacy} \\
    \midrule
    Authenticity & 1.00 & 1.00 & 1.00 & 1.00 & 1.00 & 1.00 & 1.00 & 1.00 \\
    Credibility  & 1.00 & 1.00 & 1.00 & 1.00 & 1.00 & 1.00 & 1.00 & 1.00 \\
    Matching      & 0.75 & 0.74 & 0.72 & 0.74 & 0.67 & 0.70 & 0.83 & 0.71 \\
    Fluency       & 0.72 & 0.71 & 0.69 & 0.71 & 0.64 & 0.67 & 0.80 & 0.68 \\
    Novelty      & 0.71 & 0.70 & 0.68 & 0.70 & 0.63 & 0.66 & 0.79 & 0.67 \\
    \bottomrule
    \end{tabular}
    }
    \caption{Metric evaluation results across different categories.}
    \label{tab:metric_results}
\end{table}

\section{Details of Naive Setting Prompts}
For the naive experimental settings, we also disclose its prompt in detail, see Table.\ref{tab:prompt0shot} for Naive-0-Shot, Table.\ref{tab:prompt1shot} for Naive-1-Shot, and Table.\ref{tab:promptcotshot} for Naive-Cot setting.

\section{More Cases of QUILL Application}
In this study, we conduct a comprehensive case
analysis to demonstrate the efficacy and alignment
of our reranking metric of the unsupervised UPR, the closed-source model GPT-3.5-turbo, and our approach with human evaluation.
As shown in Table.\ref{tab:appendix2}, we show the results of rag without rearrangement, index rearrangement and manual evaluation. The darker the color, the higher the manual evaluation score.

\begin{table*}[!htbp]
\centering

\resizebox{\textwidth}{!}{
\begin{tabular}{|c|cc|}
\hline
  \multirow{1}{*}{\textbf{Method}} &
  \multicolumn{1}{c}{\multirow{1}{*}{\textbf{Literal Sentence}}} &
  \multirow{1}{*}
  {\begin{tabular}[c]{|p{5cm}|p{5cm}|p{5cm}}
  \textbf{Recalled List} &
  \textbf{Metric Rerank} &
  \textbf{Human Rerank}
  
  \end{tabular}}
  \\
  \hline
  \multirow{5}{*}{GPT}
  &
  \multirow{5}{*}{\begin{tabular}[c]{@{}p{10cm}@{}}
  \begin{CJK}{UTF8}{gbsn}
康托尔在研究集合论的过程中，深刻认识到数学探索的本质。他认为：{[}Q{]}。这句话强调了在数学研究中，发现和提出新的问题比解决已有问题更为重要。康托尔的集合论突破了传统数学的界限，提出了无穷集合和基数的概念，为数学理论的发展开辟了新的道路。
\end{CJK} \end{tabular}} 
  &
  {\begin{tabular}[c]{|p{5cm}|p{5cm}|p{5cm}}
\begin{CJK}{UTF8}{gbsn} \cellcolor{green5}数学是科学的皇后，而数论是数学的皇后\end{CJK} & \begin{CJK}{UTF8}{gbsn} \cellcolor{green2}新的数学方法和概念，常常比解决数学问题本身更重要\end{CJK} & \begin{CJK}{UTF8}{gbsn} \cellcolor{green1}在数学的领域中，提出问题的艺术比解答问题的艺术更为重要\end{CJK} 
  \\
  \hline
\begin{CJK}{UTF8}{gbsn} \cellcolor{green2}新的数学方法和概念，常常比解决数学问题本身更重要\end{CJK} & \begin{CJK}{UTF8}{gbsn} \cellcolor{green1}在数学的领域中，提出问题的艺术比解答问题的艺术更为重要\end{CJK} & \begin{CJK}{UTF8}{gbsn} \cellcolor{green2}新的数学方法和概念，常常比解决数学问题本身更重要\end{CJK} 
   \\
   \hline
\begin{CJK}{UTF8}{gbsn} \cellcolor{green1}在数学的领域中，提出问题的艺术比解答问题的艺术更为重要\end{CJK} & \begin{CJK}{UTF8}{gbsn} \cellcolor{green3}数学是一种理性的精神，使人类的思维得以运用到最完善的程度\end{CJK} & \begin{CJK}{UTF8}{gbsn} \cellcolor{green3}数学是一种理性的精神，使人类的思维得以运用到最完善的程度\end{CJK}
   \\
    \hline
\begin{CJK}{UTF8}{gbsn} \cellcolor{green3}数学是一种理性的精神，使人类的思维得以运用到最完善的程度\end{CJK} & \begin{CJK}{UTF8}{gbsn} \cellcolor{green4}数学之所以有高声誉，另一个理由就是数学使得自然科学实现定理化，给予自然科学某种程度的可靠性\end{CJK} & \begin{CJK}{UTF8}{gbsn} \cellcolor{green4}数学之所以有高声誉，另一个理由就是数学使得自然科学实现定理化，给予自然科学某种程度的可靠性\end{CJK} 
  \\
  \hline
 \begin{CJK}{UTF8}{gbsn} \cellcolor{green4}数学之所以有高声誉，另一个理由就是数学使得自然科学实现定理化，给予自然科学某种程度的可靠性\end{CJK} & \begin{CJK}{UTF8}{gbsn} \cellcolor{green5}数学是科学的皇后，而数论是数学的皇后\end{CJK} & \begin{CJK}{UTF8}{gbsn} \cellcolor{green5}数学是科学的皇后，而数论是数学的皇后\end{CJK}
  \end{tabular}}
\\
\hline
  \multirow{5}{*}{Ours} &
  \multirow{5}{*}{\begin{tabular}[c]{@{}p{10cm}@{}}
  \begin{CJK}{UTF8}{gbsn}
康托尔在研究集合论的过程中，深刻认识到数学探索的本质。他认为：{[}Q{]}。这句话强调了在数学研究中，发现和提出新的问题比解决已有问题更为重要。康托尔的集合论突破了传统数学的界限，提出了无穷集合和基数的概念，为数学理论的发展开辟了新的道路。
\end{CJK}
  \end{tabular}}
  &
  {\begin{tabular}[c]{|p{5cm}|p{5cm}|p{5cm}}
\begin{CJK}{UTF8}{gbsn} \cellcolor{green5}数学是科学的皇后，而数论是数学的皇后\end{CJK} & \begin{CJK}{UTF8}{gbsn} \cellcolor{green1}礼义以生利，政事以成义\end{CJK} & \begin{CJK}{UTF8}{gbsn} \cellcolor{green1}在数学的领域中，提出问题的艺术比解答问题的艺术更为重要\end{CJK}
  \\
  \hline
\begin{CJK}{UTF8}{gbsn} \cellcolor{green2}新的数学方法和概念，常常比解决数学问题本身更重要\end{CJK} & \begin{CJK}{UTF8}{gbsn} \cellcolor{green5}数学是科学的皇后，而数论是数学的皇后\end{CJK} & \begin{CJK}{UTF8}{gbsn} \cellcolor{green2}新的数学方法和概念，常常比解决数学问题本身更重要\end{CJK}
  \\
  \hline
\begin{CJK}{UTF8}{gbsn} \cellcolor{green1}在数学的领域中，提出问题的艺术比解答问题的艺术更为重要\end{CJK} & \begin{CJK}{UTF8}{gbsn} \cellcolor{green4}数学之所以有高声誉，另一个理由就是数学使得自然科学实现定理化，给予自然科学某种程度的可靠性\end{CJK} & \begin{CJK}{UTF8}{gbsn} \cellcolor{green3}数学是一种理性的精神，使人类的思维得以运用到最完善的程度\end{CJK} 
  \\
  \hline
\begin{CJK}{UTF8}{gbsn} \cellcolor{green3}数学是一种理性的精神，使人类的思维得以运用到最完善的程度\end{CJK} & \begin{CJK}{UTF8}{gbsn} \cellcolor{green3}数学是一种理性的精神，使人类的思维得以运用到最完善的程度\end{CJK} & \begin{CJK}{UTF8}{gbsn} \cellcolor{green4}数学之所以有高声誉，另一个理由就是数学使得自然科学实现定理化，给予自然科学某种程度的可靠性\end{CJK}
  \\
  \hline
\begin{CJK}{UTF8}{gbsn} \cellcolor{green4}数学之所以有高声誉，另一个理由就是数学使得自然科学实现定理化，给予自然科学某种程度的可靠性\end{CJK} & \begin{CJK}{UTF8}{gbsn} \cellcolor{green1}在数学的领域中，提出问题的艺术比解答问题的艺术更为重要\end{CJK} & \begin{CJK}{UTF8}{gbsn} \cellcolor{green5}数学是科学的皇后，而数论是数学的皇后\end{CJK}
\end{tabular}}
\\
\hline

  \multirow{5}{*}{UPR}
  &
  \multirow{5}{*}{\begin{tabular}[c]{@{}p{10cm}@{}}
  \begin{CJK}{UTF8}{gbsn}
在荀子的政治哲学中，{[}Q{]}。是一个重要的命题。 礼义不仅是个人行为的规范，也是国家治理的基础。通过礼义，可以实现经济利益的最大化；通过公正的政务，可以实现社会正义。 荀子的这一思想在中国古代政治理论中占有重要地位，对后世的治国理政产生了深远影响。
\end{CJK} \end{tabular}} 
  &
  {\begin{tabular}[c]{|p{5cm}|p{5cm}|p{5cm}}
\begin{CJK}{UTF8}{gbsn} \cellcolor{green3}故不积跬步，无以至千里；不积小流，无以成江海\end{CJK} &
\begin{CJK}{UTF8}{gbsn}\cellcolor{green2}言无常信，行无常贞，惟利所在，无所不倾，若是则可谓小人矣\end{CJK}&
\begin{CJK}{UTF8}{gbsn}\cellcolor{green1}礼义以生利，政事以成义\end{CJK}
  \\
  \hline
\begin{CJK}{UTF8}{gbsn} \cellcolor{green5}倘能生存,我当然仍要学习\end{CJK} & \begin{CJK}{UTF8}{gbsn} \cellcolor{green1}礼义以生利，政事以成义\end{CJK} & \begin{CJK}{UTF8}{gbsn} \cellcolor{green2}言无常信，行无常贞，惟利所在，无所不倾，若是则可谓小人矣\end{CJK}
   \\
   \hline
\begin{CJK}{UTF8}{gbsn} \cellcolor{green4}玉石不经雕琢，就不能成为有用的器物；人如果不学习，就不会明白道理\end{CJK} & 
\begin{CJK}{UTF8}{gbsn} \cellcolor{green4}玉石不经雕琢，就不能成为有用的器物；人如果不学习，就不会明白道理\end{CJK} & 
\begin{CJK}{UTF8}{gbsn} \cellcolor{green3}故不积跬步，无以至千里；不积小流，无以成江海\end{CJK}
   \\
    \hline
\begin{CJK}{UTF8}{gbsn} \cellcolor{green2}言无常信，行无常贞，惟利所在，无所不倾，若是则可谓小人矣\end{CJK} & \begin{CJK}{UTF8}{gbsn} \cellcolor{green3}故不积跬步，无以至千里；不积小流，无以成江海\end{CJK} & \begin{CJK}{UTF8}{gbsn} \cellcolor{green4}玉石不经雕琢，就不能成为有用的器物；人如果不学习，就不会明白道理\end{CJK} \
   \\
  \hline
\begin{CJK}{UTF8}{gbsn} \cellcolor{green1}礼义以生利，政事以成义\end{CJK} & \begin{CJK}{UTF8}{gbsn} \cellcolor{green5}倘能生存,我当然仍要学习\end{CJK} & \begin{CJK}{UTF8}{gbsn} \cellcolor{green5}倘能生存,我当然仍要学习\end{CJK}
  \end{tabular}}
\\
\hline
  \multirow{5}{*}{Ours} &
  \multirow{5}{*}{\begin{tabular}[c]{@{}p{10cm}@{}}
  \begin{CJK}{UTF8}{gbsn}
在荀子的政治哲学中，{[}Q{]}。是一个重要的命题。 礼义不仅是个人行为的规范，也是国家治理的基础。通过礼义，可以实现经济利益的最大化；通过公正的政务，可以实现社会正义。 荀子的这一思想在中国古代政治理论中占有重要地位，对后世的治国理政产生了深远影响。
\end{CJK}
  \end{tabular}}
  &
  {\begin{tabular}[c]{|p{5cm}|p{5cm}|p{5cm}}
\begin{CJK}{UTF8}{gbsn} \cellcolor{green3}故不积跬步，无以至千里；不积小流，无以成江海\end{CJK} & \begin{CJK}{UTF8}{gbsn} \cellcolor{green1}礼义以生利，政事以成义\end{CJK} & \begin{CJK}{UTF8}{gbsn} \cellcolor{green1}礼义以生利，政事以成义\end{CJK}
  \\
  \hline
\begin{CJK}{UTF8}{gbsn} \cellcolor{green3}倘能生存,我当然仍要学习\end{CJK} & \begin{CJK}{UTF8}{gbsn} \cellcolor{green2}言无常信，行无常贞，惟利所在，无所不倾，若是则可谓小人矣\end{CJK} & \begin{CJK}{UTF8}{gbsn} \cellcolor{green2}言无常信，行无常贞，惟利所在，无所不倾，若是则可谓小人矣\end{CJK}
  \\
  \hline
\begin{CJK}{UTF8}{gbsn} \cellcolor{green4}玉石不经雕琢，就不能成为有用的器物；人如果不学习，就不会明白道理\end{CJK} & \begin{CJK}{UTF8}{gbsn} \cellcolor{green4}玉石不经雕琢，就不能成为有用的器物；人如果不学习，就不会明白道理\end{CJK} & \begin{CJK}{UTF8}{gbsn} \cellcolor{green3}故不积跬步，无以至千里；不积小流，无以成江海\end{CJK}
  \\
  \hline
\begin{CJK}{UTF8}{gbsn} \cellcolor{green2}言无常信，行无常贞，惟利所在，无所不倾，若是则可谓小人矣\end{CJK} & \begin{CJK}{UTF8}{gbsn} \cellcolor{green3}故不积跬步，无以至千里；不积小流，无以成江海\end{CJK} & \begin{CJK}{UTF8}{gbsn} \cellcolor{green4}玉石不经雕琢，就不能成为有用的器物；人如果不学习，就不会明白道理\end{CJK}
  \\
  \hline
\begin{CJK}{UTF8}{gbsn} \cellcolor{green1}礼义以生利，政事以成义\end{CJK} & \begin{CJK}{UTF8}{gbsn} \cellcolor{green5}倘能生存,我当然仍要学习\end{CJK} & \begin{CJK}{UTF8}{gbsn} \cellcolor{green5}倘能生存,我当然仍要学习\end{CJK}
\end{tabular}}
\\
\hline

\end{tabular}
}
\caption{ Additional example of recalled candidates reranked via different rerank metrics and human evaluation. 
The indicators [Q] denotes the insertion positions of the given context. A darker shade of green indicates a higher rank bestowed by humans.
}
\label{tab:appendix2}
\end{table*}

\begin{table*}[t]
\small
    \begin{tabularx}{\linewidth}{X}
    \toprule
    \color{gray}{/* \textit{Task prompt} */}\\
Suppose you are a literary scholar and are familiar with many famous people's quotes. You are required to populate contextualised quotes based on user input text within the specified [Q] symbols. \\
    \color{gray}{/* \textit{Output requirements} */}\\
1. The famous quotes must be quotes from a famous person in history or in the present, Please output the quote in English.\\
2. The quote should be closely related to the context, so that the context is more reasonable, smooth and beautiful.\\
3. If there is a specified author in the context, the famous quote must be given according to the corresponding restrictions.\\
4. Output Formate: "quote".\\
5. Only output the quote, NO MORE INFORMATION!\\
6. The number of quote should be 5 to 30 words.\\
    \color{gray}{/* \textit{Input} */}\\
    ---INPUT---\\
    \{\textbf{Query}\}\\
    ---OUTPUT---\\
    \bottomrule
    \end{tabularx}
  \caption{
    The details of the prompt for Naive-0-Shot setting.
  }
  \label{tab:prompt0shot}
\end{table*}

\begin{table*}[t]
\small
    \begin{tabularx}{\linewidth}{X}
    \toprule
    \color{gray}{/* \textit{Task prompt} */}\\
Suppose you are a literary scholar and are familiar with many famous people's quotes. You are required to populate contextualised quotes based on user input text within the specified [Q] symbols. \\
    \color{gray}{/* \textit{Output requirements} */}\\
1. The famous quotes must be quotes from a famous person in history or in the present, Please output the quote in English.\\
2. The quote should be closely related to the context, so that the context is more reasonable, smooth and beautiful.\\
3. If there is a specified author in the context, the famous quote must be given according to the corresponding restrictions.\\
4. Output Formate: "quote".\\
5. Only output the quote, NO MORE INFORMATION!\\
6. The number of quote should be 5 to 30 words.\\
\color{gray}{/* \textit{Example} */}\\
---INPUT---\\
.[Q], said by Confucius in Analects of Confucius - Wei Linggong. So is reading. Hard reading is the foundation, good reading is the key. In order to read effectively, you also need to make use of its "tools".\\
---OUTPUT---\\
"To do a good job, you must first sharpen your tools."\\
    \color{gray}{/* \textit{Input} */}\\
    ---INPUT---\\
    \{\textbf{Query}\}\\
    ---OUTPUT---\\
    \bottomrule
    \end{tabularx}
  \caption{
    The details of the prompt for Naive-1-Shot setting.
  }
  \label{tab:prompt1shot}
\end{table*}

\begin{table*}[t]
\small
    \begin{tabularx}{\linewidth}{X}
    \toprule
    \color{gray}{/* \textit{Task prompt} */}\\
Suppose you are a literary scholar and are familiar with many famous people's quotes. You are required to populate contextualised quotes based on user input text within the specified [Q] symbols. \\
    \color{gray}{/* \textit{Output requirements} */}\\
1. The famous quotes must be quotes from a famous person in history or in the present, Please output the quote in English.\\
2. The quote should be closely related to the context, so that the context is more reasonable, smooth and beautiful.\\
3. If there is a specified author in the context, the famous quote must be given according to the corresponding restrictions.\\
4. Output Formate: "quote".\\
5. Only output the quote, NO MORE INFORMATION!\\
6. The number of quote should be 5 to 30 words.\\
\\
Please think step by step then return the result!!!\\
\color{gray}{/* \textit{Examples} */}\\
1:
---INPUT---\\
.[Q], said by Confucius in Analects of Confucius - Wei Linggong. So is reading. Hard reading is the foundation, good reading is the key. In order to read effectively, you also need to make use of its "tools".\\
---OUTPUT---\\
"To do a good job, you must first sharpen your tools."\\
2:
---INPUT---\\
.[Q]. As an ancient civilisation and a responsible power, it has always been China's pursuit to help the world. By guiding the direction of the world's changing circumstances with Chinese concepts, Chinese-style modernisation will advance and expand in benign interaction with the world, and will also strengthen the power for world peace and provide opportunities for the development of all countries.\\
---OUTPUT---\\
"Already wanting to be established, we should be established; already wanting to achieve, we should achieve."\\
    \color{gray}{/* \textit{Input} */}\\
    ---INPUT---\\
    \{\textbf{Query}\}\\
    ---OUTPUT---\\
    \bottomrule
    \end{tabularx}
  \caption{
The details of the prompt for Naive-CoT setting.
  }
  \label{tab:promptcotshot}
\end{table*}

\section{nDCG Formulation}
In our experiment, in order to get the 
relevance between quote and query, we first use GPT-4o to score the relevance and get the complete 
relevance list after manual sampling. Hence, given $m$ candidate quotes $Q=\{q_1,q_2,\cdots,q_m\}$, the nDCG@k is defined as follows:
\begin{equation}
    \text{nDCG}(k) = \frac{\text{DCG}(O_\text{real}, k)}{\text{DCG}(O_\text{ideal}, k)}
\end{equation}
\begin{equation}
    \text{DCG}(O, k) = \sum_{i=1}^k \frac{Rel_i}{\log_2(1 + i)}
\end{equation}
where \( O_\text{ideal} \) and \( O_\text{real} \) represent the score list given by the ideal ranking relevance and the real ranking relevance respectively, $Rel_i$ denote the relevance score of the quote $q_i$.

\end{document}